\documentclass[journal]{IEEEtran}
\usepackage{cite}
\usepackage{graphicx}
\usepackage{dcolumn}
\usepackage{amsmath,amsfonts,amssymb}
\usepackage{subfigure}
\usepackage{bm}
\usepackage[dvips]{epsfig}
\usepackage{epsfig}
\usepackage{lettrine}
\usepackage[dvips]{graphicx}
\usepackage{multirow}
\usepackage{color}
\usepackage{float}
\usepackage{stfloats}
\usepackage{lipsum}
 \usepackage{algorithm}
\usepackage{algpseudocode}

\DeclareMathOperator*{\argmax}{argmax}
\begin{document}
\title{Robust Tensor Decomposition for Image Representation Based on Generalized Correntropy}

\author{Miaohua~Zhang,~\IEEEmembership{Student Member,~IEEE,}
        Yongsheng~Gao,~\IEEEmembership{Senior Member,~IEEE,}\\
        Changming~Sun,~\IEEEmembership{}
        and Michael~Blumenstein~\IEEEmembership{}
\thanks{This research was supported in part by Australian Research Council (ARC) under Discovery Grant DP140101075.}
\thanks{M. Zhang and Y. Gao are with Griffith School of Engineering, Griffith University, Australia (e-mails: lena.zhang@griffithuni.edu.au; yongsheng.gao@griffith.edu.au).}
\thanks{C. Sun is with CSIRO Data61, PO Box 76, Epping, NSW 1710, Australia (e-mail: changming.sun@csiro.au).}
\thanks{M. Blumenstein is with School of Software, University of Technology, Sydney, Australia (email: michael.blumenstein@uts.edu.au) }}

 \markboth{IEEE Trans. Draft}
 {Miaohua Zhang \MakeLowercase{\textit{et al.}}: Robust Tensor Decomposition for Image Representation Based on Generalized Correntropy}
 \title{Robust Tensor Decomposition for Image Representation Based on Generalized Correntropy}
  \maketitle

\begin{abstract}
Traditional tensor decomposition methods, e.g., two dimensional principal component analysis and two dimensional singular value decomposition, that minimize mean square errors, are sensitive to outliers. To overcome this problem, in this paper we propose a new robust tensor decomposition method using generalized correntropy criterion (Corr-Tensor). A Lagrange multiplier method is used to effectively optimize the generalized correntropy objective function in an iterative manner. The Corr-Tensor can effectively improve the robustness of tensor decomposition with the existence of outliers without introducing any extra computational cost. Experimental results demonstrated that the proposed method significantly reduces the reconstruction error on face reconstruction and improves the accuracies on handwritten digit recognition and facial image clustering.
\end{abstract}



\section{Introduction}
\IEEEPARstart{A}{s}
a simple and effective dimensional reduction method, principal component analysis (PCA) has been used in many computer vision applications such as face reconstruction and representation, object recognition, and tracking. However, traditional PCA methods~\cite{kwak2008principal,ding2006r,candes2011robust,Qianqianl2pnorm,rahmani2016coherence} are based on a vector space model. If original samples are given as  matrices (2D data), each matrix needs to be transformed into a vector to form a large training matrix. In this way, unfortunately, the underlying spatial (structural) information of the original data is destroyed and thus this matrix-to-vector  transformation procedure is not optimal for the extraction of the most representative features~\cite{inoue2006equivalence,ye2005two,gu2012low,hou20172d,shikkenawis20162d}. Moreover, the dimension of the vector space might be very high.

To further exploit the spatial information carried by images, many researchers in computer vision and pattern recognition considered to apply PCA methods directly on image matrices or tensors. These methods treat an image as a second-order tensor and their objective functions are expressed as functions of an image matrix instead of a high-dimensional image vector~\cite{li2010l1,yang2004two,cai2005subspace,ye2005generalized,ding20052}. Yang et al.~\cite{yang2004two} proposed a two dimensional PCA (2DPCA) method in which image covariance matrices are constructed directly using original image matrices. Cai et~al.~\cite{cai2005subspace} considered an image as a second order tensor and proposed a tensor subspace learning algorithm. Unlike 2DPCA which employs a one-sided transformation, Ye~\cite{ye2005generalized} proposed a two-sided linear transformation called the generalized low-rank approximations of matrices (GLRAM) and used an iterative procedure to solve it. Ding and Ye~\cite{ding20052} proposed a non-iterative two dimensional singular value decomposition (2DSVD) algorithm.

Although the aforementioned vector-based PCA and matrix-based PCA methods have been successfully applied to many applications, they are sensitive to outliers as they obtain the optimal solutions by minimizing an $L_2$ norm function. The effect of outliers will be exaggerated by the use of the $L_2$ norm. It is well recognized that an $L_1$ norm is more robust to outliers~\cite{kwak2008principal}\cite{ke2005robust}\cite{zhong2013linear}\cite{liu2017non} than an $L_2$ norm. Therefore, several methods have been developed to adopt the $L_1$ norm metric to measure the error between the original data and its reconstruction, e.g., $L_1$-PCA~\cite{kwak2008principal} and $L_1$-2DPCA~\cite{li2010l1}. However, these methods need to solve the optimization problem through quadratic programming that is computationally expensive and they are not rotation invariant. Then Ding et~al.~\cite{ding2006r} proposed a rotational invariant $L_1$ norm PCA ($R_1$-PCA) to improve its rotational invariance. In $R_1$ norm, the distances in spatial dimensions are measured in $L_2$ norm while the summation over different data points uses $L_1$ norm. Therefore, the $R_1$ norm reserves the rotational invariance property of $L_2$ norm and the robustness of $L_1$ norm.  In~\cite{huang2008robust}, Huang and Ding took the advantage of the $R_1$ norm and applied it to tensor decomposition, which improves the robustness of tensor decomposition methods. However, both $L_1$ and $R_1$ norm based methods require the data to be already centered, which is difficult to achieve in practice especially when there are outliers. Outliers will make the data mean biased and will then lead to a robustness decrease for the algorithms~\cite{he2011robust}.

The correntropy is a potentially promising information theoretic learning (ITL) based measurement~\cite{he2011maximum,liu2007correntropy,principe2000information,santamaria2006generalized} in handling nonzero mean and non-Gaussian noise with large outliers in the signal processing and computer vision fields. In existing subspace learning algorithms as mentioned above, mean square error (MSE) or $L_1$-norm is used in the objective function to minimize the representation error. However, MSE or $L_1$-norm cannot easily control the large errors caused by outliers because the MSE based loss functions will magnify the effect of outliers~\cite{he2011robust}~\cite{he2011maximum}\cite{chen2016generalized}\cite{chen2017robust}. Although the $L_1$-norm based loss functions minimize the loss error more than the MSE based ones, learned features are still greatly affected by outliers. The correntropy calculates the loss between the original and reconstructed data with a Gaussian function with which the correntropy induces a nonlinear metric called the correntropy induced metric. However, a Gaussian kernel is not always the best choice for maximum reduction of the loss. To address the above problems, in this paper, we propose a robust tensor factorization algorithm called Corr-Tensor which effectively improves the robustness of tensor representation by taking advantages of the generalized correntropy criterion~\cite{chen2016generalized}\cite{zhao2017kernel}. We extend the error order in the Gaussian function from the second order to an arbitrary order (denoted as $\alpha$) and thus the proposed generalized correntropy loss (Corr-Loss) function performs like different norms of data in different regions. A visual illustration of the Corr-Loss function is given in Fig. 1 where we displayed the 3D surface of the Corr-Loss function with a different error power $\alpha$.

The contributions of our work can be summarized as follows:
\begin{itemize}
\item 	To the best of our knowledge, this is the first work to explore the robustness of tensor decomposition from the perspective of a generalized correntropy with an arbitrary error power.
 \item 	We find a robust and effective representation model for tensor decomposition and it is rotationally invariant.
   \item During the optimization, the proposed method can update the data mean automatically, thus it can handle non-centered data.
  \item	We also propose a higher order tensor decomposition framework and a non-second order statistics measure in the kernel space.
 \item 	A center-based nearest neighbour classifier based on the generalized correntropy is developed to minimize the effect from outliers.
   \end{itemize}

The remainder of this paper is organized as follows: Related works are introduced in Section II. The definition of the generalized correntropy similarity measurement is introduced in Section III. The proposed method, optimization procedures and generalization of the proposed algorithm are presented in Section IV. The implementation that will be used in our experiments is introduced in Section V. In Section VI, we extended the proposed method to a generalized $p$-order form. Section VII gives experimental results of our algorithm against benchmark methods. Conclusions are drawn in Section IX.
\section{Related Works and Motivations}
Consider a set of samples $\{X_1,X_2,\ldots,X_N\}$, and each sample $X_i$ is a two dimensional image with size $a\times b$. Two dimensional approaches directly apply matrix decomposition on 2D images. Yang et al. directly used all the 2D images in a dataset to construct a covariance matrix for image representation and proposed a two dimensional PCA algorithm~\cite{yang2004two}. The covariance matrix can be calculated as:

 \begin{equation}
 C=\sum_{i=1}^N \left(X_i-\bar{X} \right)^T\left(X_i-\bar{X} \right)=\sum_{i=1}^N \hat{X}_i^T\hat{X}_i,\label{1}
 \end{equation}
 where $\bar{X}=\frac{1}{N}\sum_{i=1}^N X_i$ is the mean image of the dataset. $\hat{X}_i=(X_i -\bar{X})$ denotes the data after subtracting $\bar{X}$ from $X_i$. The optimal principal components are the orthonormal eigenvectors of $C$ corresponding to the first $k$ largest eigenvalues. The formulation in \cite{yang2004two} is actually a one-sided decomposition only by considering one dimensional column-column correlation.

 Unlike 2DPCA which only considers one-sided transformation, Ding and Ye~\cite{ding20052} proposed a 2DSVD method based on the row-row and column-column covariance matrices, in which they compute a two-sided low-rank approximation of matrices by minimizing an approximation error:
 \begin{equation}
\underset{L,M, R}{\min} J(L,M,R)=\sum_{i=1}^N\|\hat{X}_i-LM_iR^T\|_F^2,\label{2}
 \end{equation}
 where $L\in\Re^{a\times k_1}$, $M=\{M_i\}_{i=1}^N$, $R\in\Re^{b\times k_2}$, and $M_i\in\Re^{k_1\times k_2}$. The row-row and column-column covariance matrices can be defined as:

 \begin{equation}
 C_1=\sum_{i=1}^N \hat{X}_iRR^T\hat{X}_i^T, ~~~
 C_2=\sum_{i=1}^N \hat{X}_i^TLL^T\hat{X}_i.\label{3}
\end{equation}

The projection matrices $L$ and $R$ are the first $k_1$ and $k_2$ eigenvectors of $C_1$ and $C_2$, respectively.

The objective function of the 2DSVD is based on the $L_2$ norm which is likely to magnify the effect from heavy noise or outliers. Huang and Ding talked about using an $L_1$ norm based cost function to overcome this drawback in~\cite{huang2008robust}. However, $L_1$ norm based 2DSVD algorithm is computational expensive and is rotational variant. Then they proposed a robust rotational invariant 2DSVD ($R_1$-2DSVD) algorithm by taking advantages the rotational invariance property of the $L_2$ norm and the outlier resistance ability of the $L_1$ norm. The objective function of 2DSVD using $R_1$ norm is defined as:
 \begin{equation}
 \underset{L, M, R}{\min} J(L,M,R)=\sum_{i=1}^{N} (\|\hat{X}_i-LM_iR^T\|^2)^{1/2},\label{4}
\end{equation}
where $L$, $M$, and $R$ are the same size as defined in~(\ref{2}).
Different from the original 2DSVD in~(\ref{2}), the projection matrices $L$ and $R$ in $R_1$-2DSVD are computed from two reweighted covariance matrices $C_1$ and $C_2$:
 \begin{equation}
 C_1=\sum_{i=1}^N w_i~\hat{X}_iRR^T\hat{X}_i^T ,~~~C_2=\sum_{i=1}^N w_i~\hat{X}_i^TLL^T\hat{X}_i,
 \end{equation}
where $w_i=1/(\textrm{Tr}(\hat{X}_i^T\hat{X}_i-\hat{X}_i^TLL^T\hat{X}_iRR^T ))^{1/2}$.

Unfortunately, both 2DPCA and 2DSVD directly decompose tensor training samples into several projection matrices without any additional constraints for outliers, which is likely to make the learned projected matrices skewed by noise and outliers, thus their performance degrades when the level of outliers increases. Huang and Ding~\cite{huang2008robust} allocate a weight to each sample so that normal samples and outliers can be treated differently. However, the parameter to determine the weights is empirically set. Moreover, all of these methods assume that the training data are already centered, which is difficult to ensure, especially when there are outliers in the data.

To solve the above problems, we propose an algorithm based on ITL which is able to preserves the nonparametric nature of correlation learning and mean square error adaptation, but extracts more information from the data for adaptation, and yields, therefore, accurate solutions when handling non-Gaussian noise and nonlinear distributed data~\cite{liu2007correntropy}\cite{erdogmus2002generalized}\cite{hild2006feature}. The correntropy is a correlation function that is derived from ITL by extending the fundamental definition of correlation function for a random process. Compared with existing work, the correntropy based methods have the superiority in handling nonzero mean and non-Gaussian noise with large outliers. In this paper, we further extend the power of the representation error in the Gaussian kernel from the second order to an arbitrary order in the generalized correntropy measurement and proposed a Corr-Tensor algorithm so that we have more choices in controlling the error. Once the error term is flexibly controlled, the outlier samples will be easily distinguished. In the following sections, we show the details of our algorithm.

 \section{Definition of Generalized Correntropy}
The correntropy is a nonlinear and local similarity measure directly related to the probablity of how similar two random variables are in a neighborhood of the joint space controlled by the kernel bandwith. It is also closely related to the Renyi's quadratic entropy using Parzen windowing method~\cite{principe2000information}, and the correntropy contains higher-order moments of the probability density function, conducts the estimation directly from the samples, and is much simpler than conventional moment expansion. The main merit of correntropy is that the adjustable kernel size provides a practical way to choose an appropriate window size and thus provides an effective way to eliminate the detrimental effect of outliers~\cite{he2011maximum}\cite{santamaria2006generalized}.

The correntropy is a generalized similarity measure between two arbitrary random variables $A$ and $B$ defined by:
 \begin{equation}
     V_\sigma(A,B)=E[k_\sigma(A-B)],
 \end{equation}
 where $E[\cdot]$ is the expectation operator and $ k_\sigma(A-B)$ is defined as the Gaussian function $\textrm{exp}(-(A-B)^2/2\sigma^2)$
 that satisfies the Mercer's theorem~\cite{huber1981robust}. The advantage of using the kernel technique is that it can nonlinearly maps the original space to a higher dimensional space and also has a clear theoretical foundation.
 \begin{figure*}
 \centering
\begin{center}
\hspace{-0.2cm}
\vspace{0cm}
  \subfigure[]{\includegraphics[width=0.72\columnwidth]{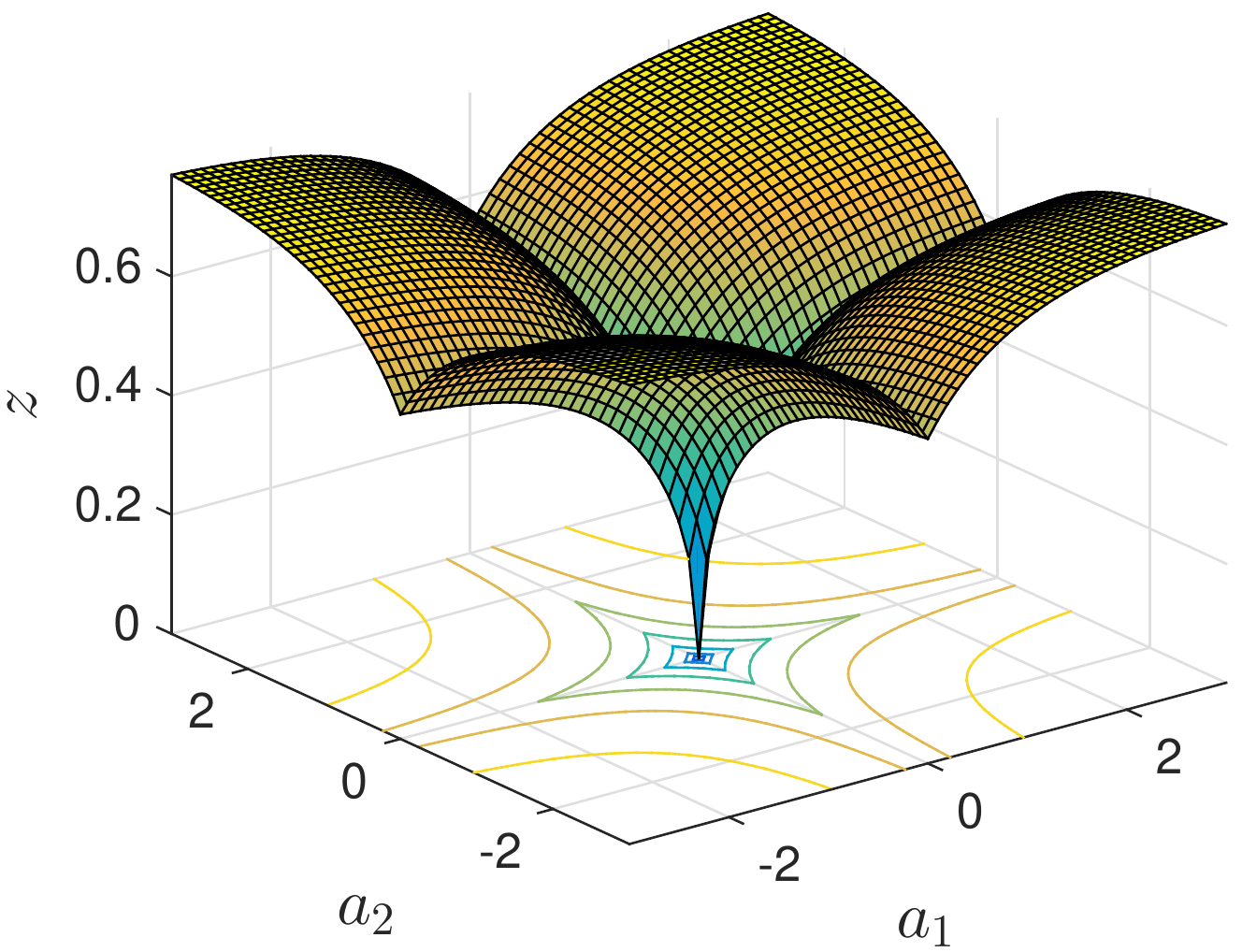}}
  \hspace{-0.7cm}
  \subfigure[]{\includegraphics[width=0.72\columnwidth]{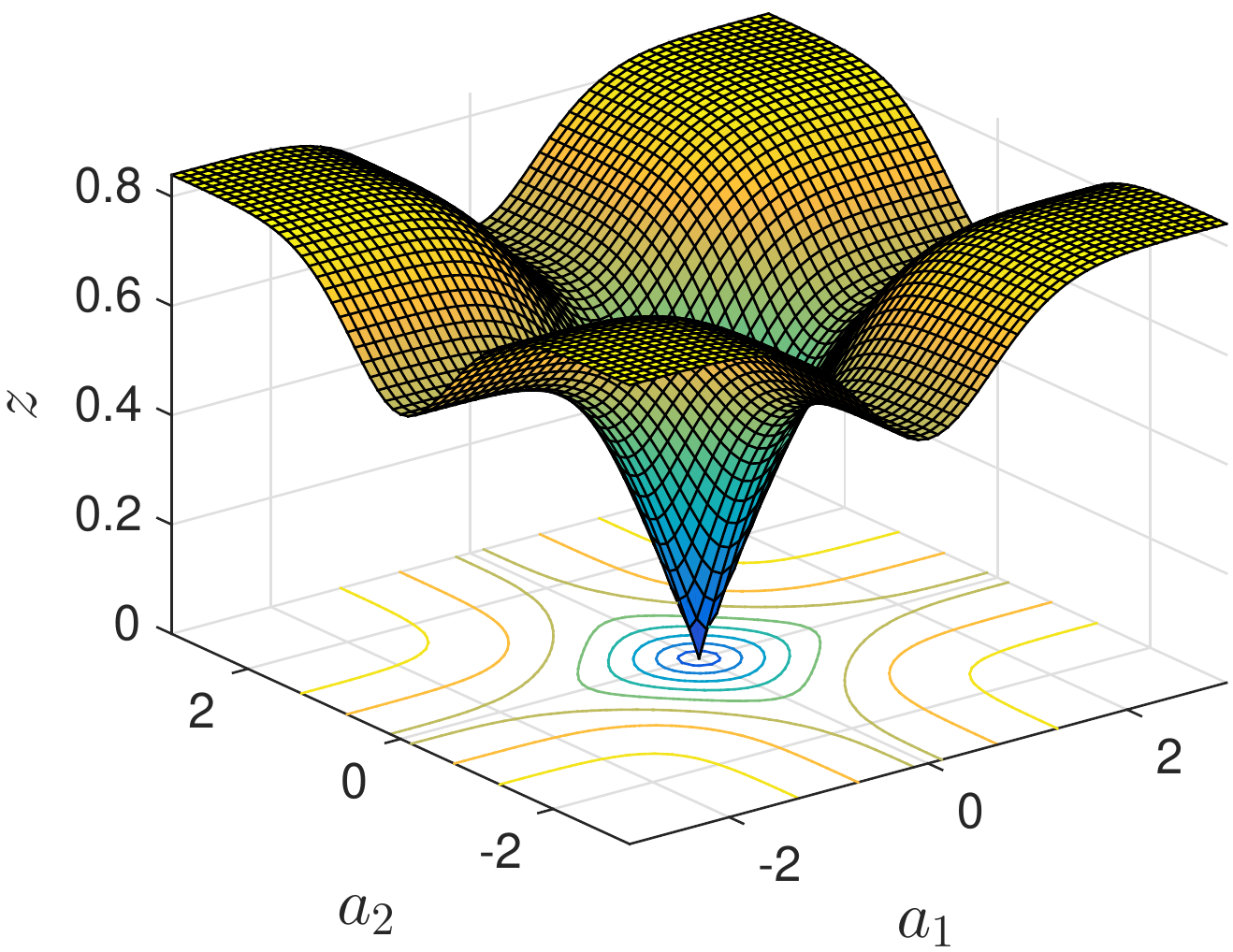}}
  \hspace{-0.7cm}
  \subfigure[]{\includegraphics[width=0.72\columnwidth]{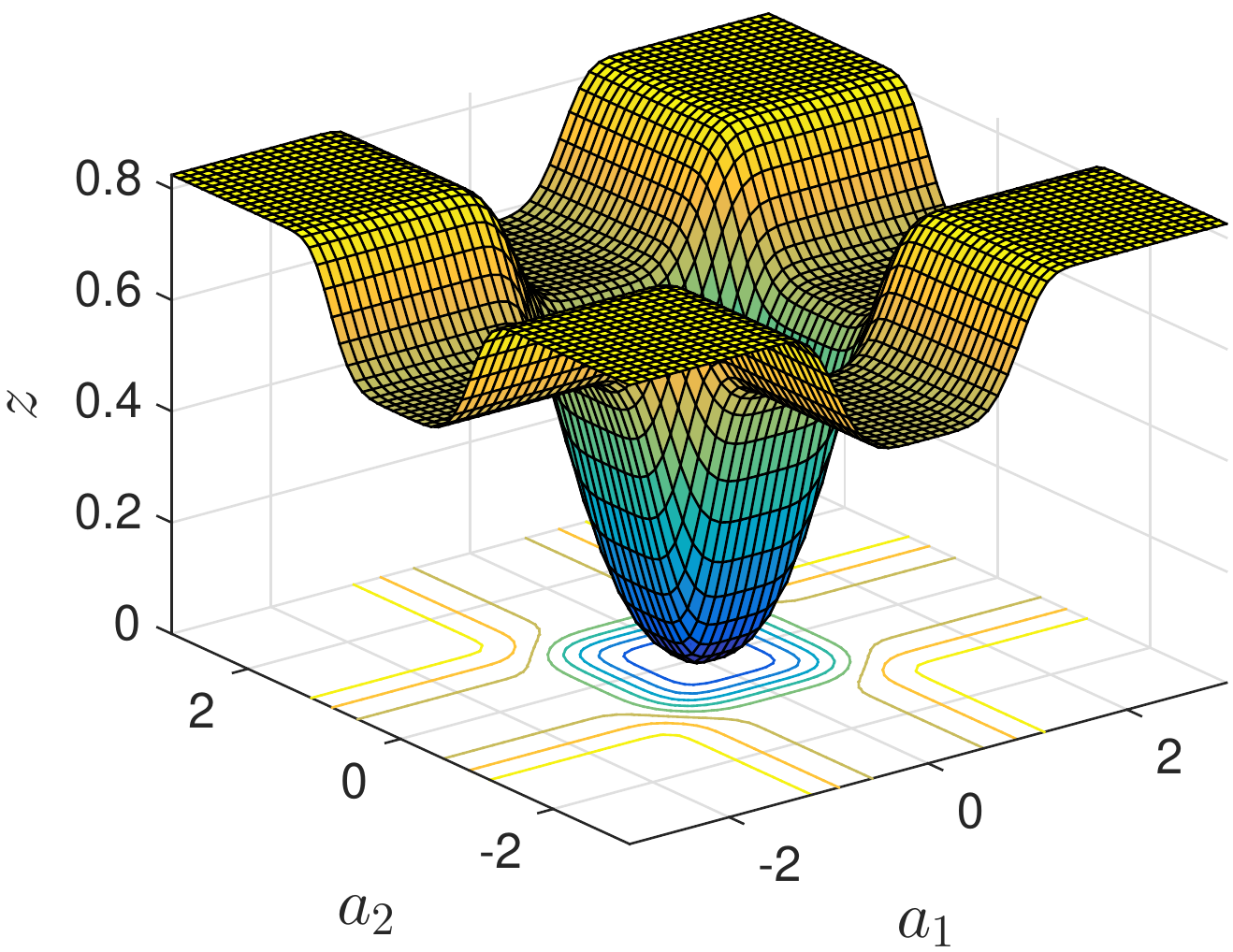}}
  \end{center}
   \caption{Surfaces of the Corr-Loss in 3D space. (a) $\alpha=1$, (b) $\alpha=2$, (c) $\alpha=5$.}
\end{figure*}

However, the kernel function of the correntropy is a Gaussian function, and we do not have much choice in controlling the contribution of error (between two random variables) to the learned features as the power of error is fixed at a quadratic form. To overcome this problem, the generalized Gaussian density (GGD) function is considered in this paper.

The GGD with zero mean is given by:
 \begin{equation}
  \begin{aligned}
  &~ G_{\alpha,\beta}(e)=\frac{\alpha}{2\beta\Gamma(1/\alpha)}\text{exp}\left(-\left|\frac{e}{\beta}\right|^\alpha\right)\\
  &~~~~~~~~~~=\gamma_{\alpha,\beta}\text{exp}\left(-\lambda\left|e\right|^\alpha\right),
  \end{aligned}
  \end{equation}
where $\alpha>0$ and $\beta>0$ are the parameters of GGD indicating the peak and width of the probability density function. $\Gamma(z)=\int_0^\infty e^{-t}t^{z-1}dt, (z>0)$ is the gamma function. $\lambda=1/\beta^\alpha$ and $\gamma_{\alpha,\beta}=\alpha/(2\beta\Gamma(1/\alpha))$ are the kernel parameter and the normalization constant, respectively. Obviously, the Gaussian function is just a special case of the generalized Gaussian density function when $\alpha$ is $2$.

To define a generalized correntropy, we use the generalized Gaussian density function as the kernel function of correntropy:

 \begin{equation}
 V_{\alpha,\beta}(A,B)=E[\varphi_{\alpha,\beta}(A)^T\varphi_{\alpha,\beta}(B)]=E[G_{\alpha,\beta}(A-B)],
 \end{equation}
where $\varphi_{\alpha,\beta}(\cdot)$ denotes a nonlinear mapping which transforms its argument into a high-dimensional Hilbert space~\cite{liu2007correntropy}.
A generalized correntropy loss function (Corr-Loss) between $A$ and $B$, which can be viewed as the second order statistical measure in the kernel space, is defined in this paper as follows:
 \begin{equation}
\begin{aligned}
&J_{\text{Corr-Loss}}(A,B)=\frac{1}{2}E\left[||\varphi_{\alpha,\beta}(A)-\varphi_{\alpha,\beta}(B)||_{\mathcal{H}}^2\right]\\
&=\frac{1}{2}E\left[\langle\varphi_{\alpha,\beta}(A)-\varphi_{\alpha,\beta}(B),\varphi_{\alpha,\beta}(A)-\varphi_{\alpha,\beta}(B)\rangle_{\mathcal{H}}\right]\\
&=\frac{1}{2}E[\langle\varphi_{\alpha,\beta}(A),\varphi_{\alpha,\beta}(A)\rangle+\langle\varphi_{\alpha,\beta}(B),\varphi_{\alpha,\beta}(B)\rangle \\
&~~~~-2\langle\varphi_{\alpha,\beta}(A),\varphi_{\alpha,\beta}(B)\rangle]\\
&=E[(G_{\alpha,\beta}(0)-G_{\alpha,\beta}(A-B))]\\
&=G_{\alpha, \beta}(0)-V_{\alpha,\beta}(A,B).
\end{aligned}
\end{equation}

However, the joint probability density function of $A$ and $B$ is often unknown in practice, thus it is difficult to compute the aforementioned expectation. Therefore, in actual implementation, the generalized correntropy is estimated by applying the Parzen windowing method to a finite number of samples $\{(a_i,b_i)\}_{i=1}^N$ available\cite{liu2007correntropy}\cite{principe2000information}\cite{santamaria2006generalized}:

 \begin{equation}
 \hat{V}_{\alpha,\beta}(A,B)=\frac{1}{N} \sum_{i=1}^N G_{\alpha,\beta}(a_i-b_i).
 \end{equation}

Thus, the Corr-Loss estimator with a set of samples $\{(a_i,b_i)\}_{i=1}^N$ can be accordingly described as follows:

\begin{equation}
\begin{aligned}
&J_{\text{Corr-Loss}}(A,B)=G_{\alpha, \beta}(0)-V_{\alpha,\beta} (A,B)\\
&=\gamma_{\alpha,\beta}-\frac{1}{N}\sum_{i=1}^N G_{\alpha, \beta}(a_i-b_i)\\
&=\gamma_{\alpha,\beta}-\frac{1}{N}\sum_{i=1}^N G_{\alpha,\beta}(e_i).
\end{aligned}
\end{equation}

To clearly describe the difference of the loss function under different $\alpha$, we plot the surfaces of the $J_{\text{Corr-Loss}}(A,B)$ in Fig. 1 with $\alpha=\{1,2,5\}$, $A=[a_1,a_2],B=[0,0], a_1\in[-3, 3], a_2\in[-3, 3]$, and $\beta=0.8$.
 \section{Corr-2DSVD}
 \subsection{Objective Function}
 Motivated by the robustness of Corr-Loss for outlier rejection, we introduce Corr-Loss into the aforementioned rotational 2DSVD method~\cite{ding2006r}.

The proposed robust vision of 2DSVD based on Corr-Loss (Corr-2DSVD) is described as follows:
 \begin{equation}
 \begin{aligned}
&\underset{L,R,\{M_i\}}{\min}~ J_{\text{Corr-Loss}}(L,R,\{M_i\},\bar{X})\\
&=\gamma_{\alpha,\beta}\left\{1-E[\text{exp}(-\lambda|e|^\alpha)]\right\}\\
& =\gamma_{\alpha,\beta}-\gamma_{\alpha,\beta}E[\text{exp}\left(-\lambda\left|\sqrt[]{||\hat{X}-LMR^T||^2}\right|^\alpha\right)].
\end{aligned}
 \end{equation}

First, we solve the optimization problem on the matrix $M_i$ by setting the derivative of $J_{\text{Corr-Loss}}$ with respect to $M_i$ to zero:
  \begin{equation}
   \begin{aligned}
 &\frac{\partial J_{\text{Corr-Loss}}}{\partial M_i}\\
  &=-\gamma_{\alpha,\beta}\lambda\alpha\text{exp}(-\lambda|e_i|^{\frac{\alpha}{2}})|e_i|^{\frac{\alpha}{2}-1}(\hat{X}_i-LM_iR^T)L^TR,
 \end{aligned}
  \end{equation}
where $e_i=||\hat{X}_i-LM_iR^T||_F^2$. Since the exponential term is not possible to be zero, we have $\hat{X}_i-LM_iR^T=0$. Then we obtain:
 \begin{equation}
  M_i=L^T\hat{X}_i R.
  \end{equation}

After replacing $M_i$ in (12) with (14), the Corr-2DSVD becomes:
\begin{equation}
\begin{aligned}
&\underset{L,R,\{M_i\}}{\min}~ J_{\text{Corr-Loss}}(L,R,\{M_i\},\bar{X})\\
& =\gamma_{\alpha,\beta}(1-\sum_{i=1}^N E[\text{exp} (-\lambda|e_i|^{\frac{\alpha}{2}})])\\
& ~~~\text{s.t.}~~~~~ L^TL=I,~~~~~ R^TR=I~,
 \end{aligned}
\end{equation}
where $e_i=\text{Tr}(\hat{X}_i^T\hat{X}_i-\hat{X}_i^TLL^T\hat{X}_i RR^T)$.
\subsection{Optimal Solutions}
Optimal solutions we need to solve are the mean matrix $\mu$, and the left and right projection matrices $L$ and $R$. Since there are two equality constraints in (15), we consider using the Lagrange multipliers method to solve the $J_{\text{Corr-Loss}}$ function with these constraints.

The Lagrange function for (15) is given by:
\begin{equation}
 \begin{aligned}
 &\mathcal{L}(\bar{X}^{t+1},L^{t+1},R^{t+1})\\
 &=J_{\text{Corr-Loss}}(\bar{X},L,R)+\text{Tr}(\Sigma(L^TL-I))+\text{Tr}(\Omega(R^TR-I)),
 \end{aligned}
 \end{equation}
where $\Sigma$ and $\Omega$ are symmetric Lagrangian multipliers. The gradient of $\mathcal{L}$ with respect to the variables (optimal solutions) must be zero. So we have:
\begin{equation}
 \begin{aligned}
 &\frac{\partial \mathcal{L}}{\partial \bar{X}}=\rho\sum_{i=1}^N \omega_i^{t+1} (\bar{X}-X_i+LL^TX_iRR^T-LL^T\bar{X} RR^T)\\
&~~~~~=0,\\
\end{aligned}
\end{equation}
where $\rho=\gamma\lambda\alpha$ and $\omega_i^{t+1}=\text{exp}(-\lambda|e_i^t|^{\frac{\alpha}{2}})|e_i^t|^{\frac{\alpha}{2}-1}$. After some algebraic operations, the optimal solution $\bar{X}^{t+1}$ can be updated by:
\begin{equation}
\bar{X}^{t+1}=\frac{\sum_{i=1}^N \omega_i^{t+1} X_i}{\sum_{i=1}^N \omega_i^{t+1}}.
\end{equation}

By taking the derivative of $\mathcal{L}$ with respect to $L$,
\begin{equation}
\begin{aligned}
&~~~~~~\frac{\partial \mathcal{L}}{\partial L}=C_1 L+L\Sigma=0\\
&\text{s.t.}~~~C_1=\frac{-\omega_i^{t+1}}{2}\rho \hat{X}_i^{t+1}R^t(R^t)^T(\hat{X}_i^{t+1})^T,\\
&~~~~~~~\rho=\gamma\lambda\alpha,\\
&~~~~~~~\omega_i^{t+1}=\text{exp}(-\lambda|e_i^{t+1}|^{\frac{\alpha}{2}})|e_i^{t+1}|^{\frac{\alpha}{2}-1},
\end{aligned}
\end{equation}
the optimal solution for $L$ is the first $k_1$ eigenvectors of $C_1$.

By taking the derivative of $\mathcal{L}$ with respect to $R$,
\begin{equation}
\begin{aligned}
&~~~~~~~~~~\frac{\partial \mathcal{L}}{\partial R}=C_2 R+R\Omega=0\\
&~~~~\text{s.t.}~~~C_2=\frac{-\omega_i^{t+1}}{2}\rho(\hat{X}_i^{t+1})^TL^{t+1}(L^{t+1})^T \hat{X}_i^{t+1}.\\
&~~~~~~~~~~~\rho=\gamma\lambda\alpha,\\
&~~~~~~~~~~~\omega_i^{t+1}=\text{exp}(-\lambda|e_i^{t+1}|^{\frac{\alpha}{2}})|e_i^{t+1}|^{\frac{\alpha}{2}-1},
\end{aligned}
\end{equation}
the optimal solution $R$ is the first $k_2$ eigenvectors of $C_2$.

Based on the above analysis, the proposed algorithm can be summarized in Algorithm 1.
\begin{algorithm}[htb]
\caption{Corr-2DSVD Algorithm}
\begin{algorithmic}[1]
\Require
   Projection matrices $L$ and $R$ calculated from 2DSVD algorithm, stopping threshold value $\epsilon$.
\Ensure
 $\{\omega_i\}_{i=1}^N$, $\bar{X}$, $L$ and $R$.
 \While {$t$=1,\dots,$T$}
\State Calculate weights $\{\omega_i\}_{i=1}^N$ using (19) and (20)
\State Update the data mean $\bar{X}^{t+1}$ using (18)
\State Update $L$ and $R$ using (19) and (20)
\State 1) using the current $L$ and $R$ to calculate $C_1$, the \indent optimal $L$ is the first $k_1$ left singular vector of $C_1$,
\State 2) using the current $L$ and $R$ to calculate $C_2$, the \indent optimal $R$ is the first $k_2$ left singular vector of $C_2$,
\If {$\epsilon>1e-5$}
\State repeat;
\Else
\State $t\leftarrow t+1$; Break;
\EndIf
\EndWhile
\end{algorithmic}
\end{algorithm}
\subsection{Generalization}
Here we extend the proposed algorithm to a higher order tensor decomposition. The input data of $N$-dimensional tensor $\mathcal{X}=\{\mathcal{X}_{i_1i_2\dots i_N}\} ~\text{with}~  i_1=1,\dots,N_1;i_2=1,\dots,N_2;\dots;i_N=1,\dots,N_n$ can be viewed as $\mathcal{X}=\{\mathcal{X}_{N_1},\mathcal{X}_{N_2},\dots,\mathcal{X}_{N_n}\}$ where each $\mathcal{X}_i$
is an $(N-1)$-dimensional tensor. We compress ($N$-1) dimensions of each tensor $\mathcal{X}_{N_i}$  but not on the data index dimension~\cite{ding2006r}. The robust version of $N$-1 tensor factorization using $R_1$ norm is:
\begin{equation}
\begin{aligned}
&\underset{\{U_n\}_{n=1}^{N-1}}{\text{min}}J=\sum_{i_N=1}^{N_n}\sqrt[]{\|\hat{\mathcal{X}}_{i_N}-U_1\otimes_1U_2 \dots U_{N-1}\otimes_{N-1}\mathcal{M}_{i_N}\|^2}\\
&~~~~~~~~~\text{s.t.}~~~~ U_n^TU_n=I,~~~~n=1,\dots,N-1,
\end{aligned}
\end{equation}
where $\hat{\mathcal{X}}_{i_N}=\mathcal{X}_{i_N}-\bar{\mathcal{X}}_{i_N}$, $\bar{\mathcal{X}}_{i_N}=\frac{1}{N_n}\sum_{{i_N}=1}^{N_n}\mathcal{X}_{i_N}$. $U\otimes_n\mathcal{M}$ denotes the $n$-mode tensor product of matrix $U$ and tensor $\mathcal{M}$.

The proposed tensor factorization using the Corr-Loss can be formulated as follows:
\begin{equation}
\begin{aligned}
&\underset{\{U_n\}_{n=1}^{N-1},\mathcal{M}_{i_N},\bar{\mathcal{X}}_{i_N}}{\min}~J_{\text{Corr-Loss}}(\{U_n\},\mathcal{M}_{i_N},\bar{\mathcal{X}}_{i_N})\\
&=\gamma_{\alpha,\beta}(1-\frac{1}{N_n}\sum_{{i_N}=1}^{N_n}\text{exp}(-\lambda|e_{i_N}|^{\frac{\alpha}{2}})),\\
&\text{s.t.}~~~~~ U_n^TU_n=I ~~~~n=1,\dots,N-1,~~~~
\end{aligned}
\end{equation}

\noindent where $e_{i_N}=\hat{\mathcal{X}}_{i_N}-U_1\otimes_1U_2\dots U_{N-1}\otimes_{N-1}\mathcal{M}_{i_N}$,
The Lagrange function for (22) is given by:
\begin{equation}
\begin{aligned}
&\mathcal{L}(\bar{\mathcal{X}},\{U_n\},\mathcal{M}_{i_N})\\
&=J_{\text{Corr-Loss}}+\sum_{n=1}^{N-1}\text{Tr}(\Sigma_n(U_n^TU_n-I)),
\end{aligned}
\end{equation}
where $\text{Tr}$ denotes the matrix trace and $\{\Sigma_n\}_{n=1}^{N-1}$ are symmetric Lagrangian multipliers. The gradient of $\mathcal{L}$ with respect to the optimal solutions $\{U_n\}_{n=1}^{N-1}$ must be zero:
\begin{equation}
\begin{aligned}
\frac{\partial \mathcal{L}}{\partial \bar{\mathcal{X}}}=\rho\sum_{{i_N}=1}^{N_n} \omega_{i_N}((1-\delta)\mathcal{X}_{i_N}+(1-\delta)\bar{\mathcal{X}})=0,
\end{aligned}
\end{equation}
where $\rho=-\gamma_{\alpha,\beta}\lambda\alpha$, $\delta=\displaystyle\prod_{n=1}^{N-1}(U_nU_n^T)$ and $\omega_{i_N}=\text{exp}(-\lambda|e_{i_N}|^{\frac{\alpha}{2}})(|e_{i_N}|^{\frac{\alpha}{2}-1})$. Thus the mean tensor $\bar{\mathcal{X}}$ can be updated by:
\begin{equation}
\bar{\mathcal{X}}=\frac{ \sum_{{i_N}=1}^{N_n} \omega_{i_N} \mathcal{X}_{i_N}}{\sum_{{i_N}=1}^{N_n} \omega_{i_N}}.
\end{equation}

Then the projection matrices $\{U_n\}_{n=1}^{N-1}$ can be updated by:
\begin{equation}
\begin{aligned}
&\frac{\partial \mathcal{L}}{\partial U_n}\\
&=\rho\sum_{{i_N}=1}^{N_n}\omega_{i_N}\sum_{i_{-n}}(\hat{\mathcal{X}}_{i_1,\dots,i_{N-1}}^{i_N}\hat{\mathcal{X}}_{i_1^{'},\dots,i_{N-1}^{'}}^{i_N} W_{-n})U_n\\
&+2\Sigma_n U_n=0,\\
&\Rightarrow CU_n=\Sigma_n U_n,
\end{aligned}
\end{equation}
where $\omega_{i_N}=\text{exp}(-\lambda|e_{i_N}|^{\frac{\alpha}{2}})|e_{i_N}|^{\frac{\alpha}{2}}$, $\rho=-\frac{\gamma_{\alpha,\beta}}{2}\lambda \alpha$, $W_{-n}=(U_1U_1^T)_{i_1 i_1^{'}} \dots (U_{n-1}U_{n-1}^T)_{i_{n-1}i_{n-1}^{'}}(U_{n+1}U_{n+1}^T)_{i_{n+1}i_{n+1}^{'}}$, $i_{-n}=i_1i_1^{'},\ldots,i_{n-1}i_{n-1}^{'}i_{n
+1}i_{n+1}^{'},\ldots,i_{N-1}i_{N-1}^{'}$, and $i_ni_n^{'}$ denotes the index of matrix $U_n$. Thus $U_n$ can be obtained by solving the eigenvectors of $C$.

\section{Implementations}
In previous sections, we have presented the proposed Corr-2DSVD algorithm to learn good features in the presence of outliers. Here we show how to apply the learned features to real image processing tasks for image reconstruction, image classification, and image clustering.
\subsection{Image Reconstruction}
For image reconstruction, denote $X_i^{\text{org}}$ as the $i$th original image of training data and $X_i^{\text{new}}$ as the reconstructed image corresponding to the original image. The aims of image reconstruction is to minimize the reconstruction error as much as possible, which can be formulated as follows:
\begin{equation}
\begin{aligned}
\min \frac{1}{N} \sum_{i=1}^N ||X_i^{\text{org}}-X_i^{\text{new}}||^2,
\end{aligned}
\end{equation}
where $X_i^{\text{new}}=L^tM_i^t(R^t)^T$, $t$ stands for solutions after the $t$th iteration.
\subsection{Image Classification}
 We aim to classify an unseen sample $X_{\text{test}}$ which comes from an existing class. Note that the proposed algorithm in Algorithm 1 aims to decompose the given training sample sets to several projection matrices in the presence of outliers. Ideally, after dimensional reduction, projected training samples from the same class of $X_{\text{test}}$ should have the smallest distance. This gives us the motivation to design a class specific nearest neighbor classifier similar to the classifier proposed by Gao and Wang in~\cite{gao2007center}, but the major difference is that the classifier introduced here is based on the generalized correntropy. Specifically, we implemented a center-based nearest neighbor classifier. The distance between two images $X_i$ and $X_j$ in tensor subspace is usually computed as~\cite{ding2008tensor}:
\begin{equation}
||X_i-X_j||_2=||LM_iR^T-LM_jR^T||_2=||M_i-M_j||_2
\end{equation}

The distance between images is thus transformed into that between matrices $M_i$ and $M_j$ of training and testing samples. The center of $M_i$ for a class $c$ is obtained by:
\begin{equation}
\begin{aligned}
&\bar{M}_c=\frac{\sum_{i\in c}^{N_c} \omega(M_i)L^TX_iR}{\sum_{i\in c}^{N_c}\omega(M_i)},
\end{aligned}
\end{equation}

\noindent where $\omega$ has the same definition as (17)-(20), namely
$\omega(M_i)=\text{exp}(-\lambda|e_i|^{\frac{\alpha}{2}})|e_i|^{\frac{\alpha}{2}-1}$.

Based on (28)-(29), the similarity measurement between the testing image and an arbitrary class center of training images based on generalized correntropy can be described as follows:
\begin{equation}
\begin{aligned}
& d(M_{\text{test}},\bar{M}_c)=G_{\alpha,\beta}(M_{\text{test}}-\bar{M_c})\\
&~~~~~~~~~~~~~~~~= \gamma_{\alpha,\beta} \text{exp}(-\lambda\sqrt[]{|M_{\text{test}}-\bar{M}_c|^{\alpha}}).
\end{aligned}
\end{equation}

The testing image is classified to the class with the maximum distance:
\begin{equation}
\underset {c}{\max}~~\{d_1,d_2,\dots,d_c, \dots,d_C\},
\end{equation}
where $C$ is the number of classes.

Algorithm 2 summarizes the classifier for Corr-2DSVD.
\begin{algorithm}[htb]
\caption{Classifier of Corr-2DSVD}
\begin{algorithmic}
\Require
  Data matrix $X=[X_1,X_2,\dots,X_N]$ from $C$ classes; a test sample $X_{\text{test}}$
\Ensure
 label($X_{\text{test}}$)
 \State 1) Calculate the center of $M$ for each class using (29)
 \State 2) Calculate the distance between the core matrix $M$ of the test image and the center of each class using (30), for $c=1,2,\dots,C$
 \State 3) $\text{label}(X_{\text{test}})=\underset{c}{\mathop{\argmax}} ~~d_c(X_{\text{test}})$
\end{algorithmic}
 \end{algorithm}

\subsection{Image Clustering}
 As mentioned in (28), the distance relationship between images can be directly quantized by that between matrix $M$ corresponding to each image, thus the tensor clustering can be carried out entirely on $M$~\cite{huang2008simultaneous}. Assuming that there are $K$ different classes, we need to cluster all the samples into $K$ clusters:
\begin{equation}
\begin{aligned}
\underset {C_k}{\min} \sum_{i=1}^N \underset {1\leq k\leq K}{\min} ||M_i-C_k||^2 =\sum_{k=1}^K \sum_{i\in C_k}||M_k^i-C_k||^2,
\end{aligned}
\end{equation}
where $C_k$ is the centroid tensor of cluster $k$.

The clustering performance is evaluated by comparing the obtained label of each sample with that provided by the dataset. Two evaluation metrics are adopted to measure the performance of clustering: the accuracy (AC) and the normalized mutual information (NMI) metric~\cite{CHH05}\cite{chen2017robust}. Denote $p_i$ and $q_i$ as the corresponding ground truth label and clustering result label of any data sample $X_i$. Then the accuracy is defined as follows:

\begin{equation}
\text{AC}=\frac{\sum_{i=1}^n \delta(p_i, \text{map}(q_i))}{n}~,
\end{equation}
where $n$ is the total number of samples, $\delta(x,y)=1$ if $x=y$, and $\delta(x,y)=0$ otherwise. $\text{map}(q_i)$ is the best mapping function which uses the Kuhn-MunKres algorithm\cite{lovasz2009matching} to permute clustering labels to match the ground truth labels.

The NMI provides a sound indication of the shared mutual information between a pair of cluster~\cite{strehl2002cluster}. Given data sample $X_i$, let $S$ and $T$ be the set of clusters obtained from the ground truth and our algorithm, respectively. The NMI is defined as follows:

\begin{equation}
\text{NMI}=\frac{I(S;T)}{[H(S)+H(T)]/2},
\end{equation}

\noindent where $I(S;T)$ is the mutual information of $S$ and $T$. $H(S)$ and $H(T)$ are the entropies of $S$ and $T$, respectively. $\text{NMI}(S,T)$ ranges from $0$ to $1$. \text{NMI} equals 1 if the two sets of clusters are identical, and NMI equals 0 if the two sets are independent. The mutual information of $S$ and $T$ can be defined as follows:
\begin{equation}
I(S,T)=\sum_{s_i\in S, t_j\in T} p(s_i,t_j) \text{log}_2\frac{p(s_i,t_j)}{p(s_i)\cdot p(t_j)},
\end{equation}
where $p(s_i)$ and $p(t_j)$ are the probabilities that a sample arbitrarily selected from the dataset belongs to clusters $s_i$ and $t_j$, respectively, and $p(s_i,t_j)$ is the joint probability that the arbitrarily selected sample belongs to the clusters $s_i$ and $t_j$ at the same time.

\section{$p$-order Extension}
The proposed algorithms that we presented above are  based on second order statistics in the kernel space. Motivated by the non-second order statistic measure which has the advantages in improving the robustness of subspace learning algorithm, in this section, we extend the proposed framework into a general version with an arbitrary order on the loss function. The Corr-Loss function with arbitrary order $p$ (Corr-PLoss) is shown as follows:
\begin{equation}
\begin{aligned}
&J_{\text{Corr-PLoss}}(X,Y)=2^{-p/2}E\left[||\varphi_{\alpha,\beta}(X)-\varphi_{\alpha,\beta}(Y)||_{\mathcal{H}}^p\right]\\
&=2^{-p/2}E\left[\left(||\varphi_{\alpha,\beta}(X)-\varphi_{\alpha,\beta}(Y)||_{\mathcal{H}}^2\right)^{p/2}\right]\\
&=2^{-p/2}E\left[\left(2G_{\alpha,\beta}(0)-2\hat{V}_{\alpha,\beta}(X,Y)\right)^{p/2}\right]\\
&=E\left[(\gamma_{\alpha,\beta}-G_{\alpha,\beta}(X-Y))^{p/2}\right],
\end{aligned}
\end{equation}
where $p>0$ is the power parameter. Obviously, the above equation (36) includes the case for the Corr-Loss function in (9) when $p$ is 2. Accordingly, by adopting a similar optimization procedure, we can derive the non-second order Corr-2DSVD algorithm and the higher order Corr-Tensor algorithm.
\section{Experimental Results}
In this section, we present experiments on publicly available databases, namely YALE\footnotemark\footnotetext[1]{http://cvc.cs.yale.edu/cvc/projects/yalefaces/yalefaces.html}, LFW \footnotemark\footnotetext[2]{http://conradsanderson.id.au/lfwcrop/},  MNIST\footnotemark\footnotetext[3]{http://yann.lecun.com/exdb/mnist/}, and ORL\footnotemark\footnotetext[4]{http://www.cl.cam.ac.uk/research/dtg/attarchive/facedatabase.html} databases, for several image processing tasks, which serve both to demonstrate the efficacy of the proposed Corr-2DSVD algorithm and to validate the claims of the previous sections. Image reconstruction, image classification, and image clustering are implemented successively to examine the quality of the learned subspaces using our framework, comparing performance across various evaluation measurements, and comparing it to several methods including 2DPCA~\cite{yang2004two}, $L_1$-2DPCA~\cite{li2010l1}, 2DSVD~\cite{ding20052}, and $R_1$-2DSVD~\cite{huang2008robust}, N-2DNPP~\cite{zhang2017robust}, and S-2DNPP~\cite{zhang2017robust}. The code of our algorithm and all the benchmarks will be available on our lab homepage\footnotemark\footnotetext[5]{https://maxwell.ict.griffith.edu.au/cvipl/publications.html}.
\subsection{Databases}
\subsubsection{YALE Face Database}The Yale face database consists
of 165 gray scale images in GIF format from 15 individuals. There are 11
images per subject, with variations of facial expressions or different
configurations. The original images are in 256 gray scales. In our experiments, all images are normalized to the range of [0,1].
\subsubsection{LFW Face Database} The LFW face database contains images of 5,749 different individuals. In this paper, we use the cropped version$^2$. There are two versions: grayscale version and color version. For each version, there are 13,233 faces and each image is resized to $64 \times 64$ (gray) or $64 \times 64 \times 3$ (color).
\subsubsection{MNIST Handwritten Digit Database}The MNIST database of handwritten digits has $60, 000$ training samples and $10,000$ testing samples. The digits have been centered in fixed-size $(28\times28)$ images whose pixels are normalized to $1$. We randomly selected $\{500,1,000,2,000,3,000\}$ images per digit in the training set as our training samples, and used all the testing samples for testing.
 \subsubsection{ORL Face Database}The ORL database has ten different images for each of the 40 distinct subjects. The images were taken at different times, with varying lighting, facial expressions (open/closed eyes, smiling/not smiling), and facial details (glasses/no glasses). All the images were taken against a dark homogeneous background with the subjects in an upright, frontal position (with tolerance for some side movement).\\\\

\begin{figure*}
\begin{center}
\hspace{-0.3cm}
\vspace{-0.5cm}
   \subfigure[]{\includegraphics[width=1\columnwidth]{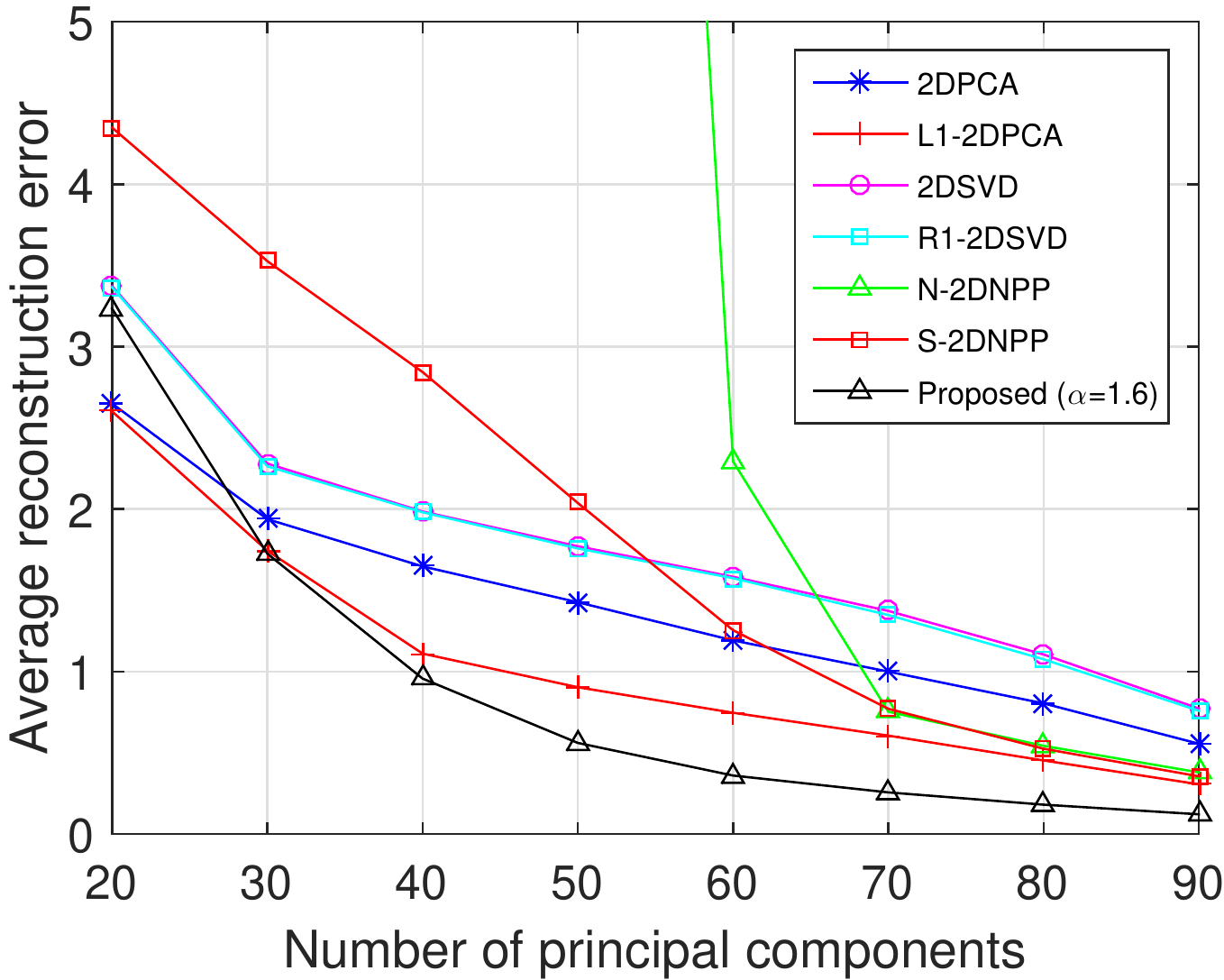}}
    \hspace{-0.2cm}
  \subfigure[]{\includegraphics[width=1\columnwidth]{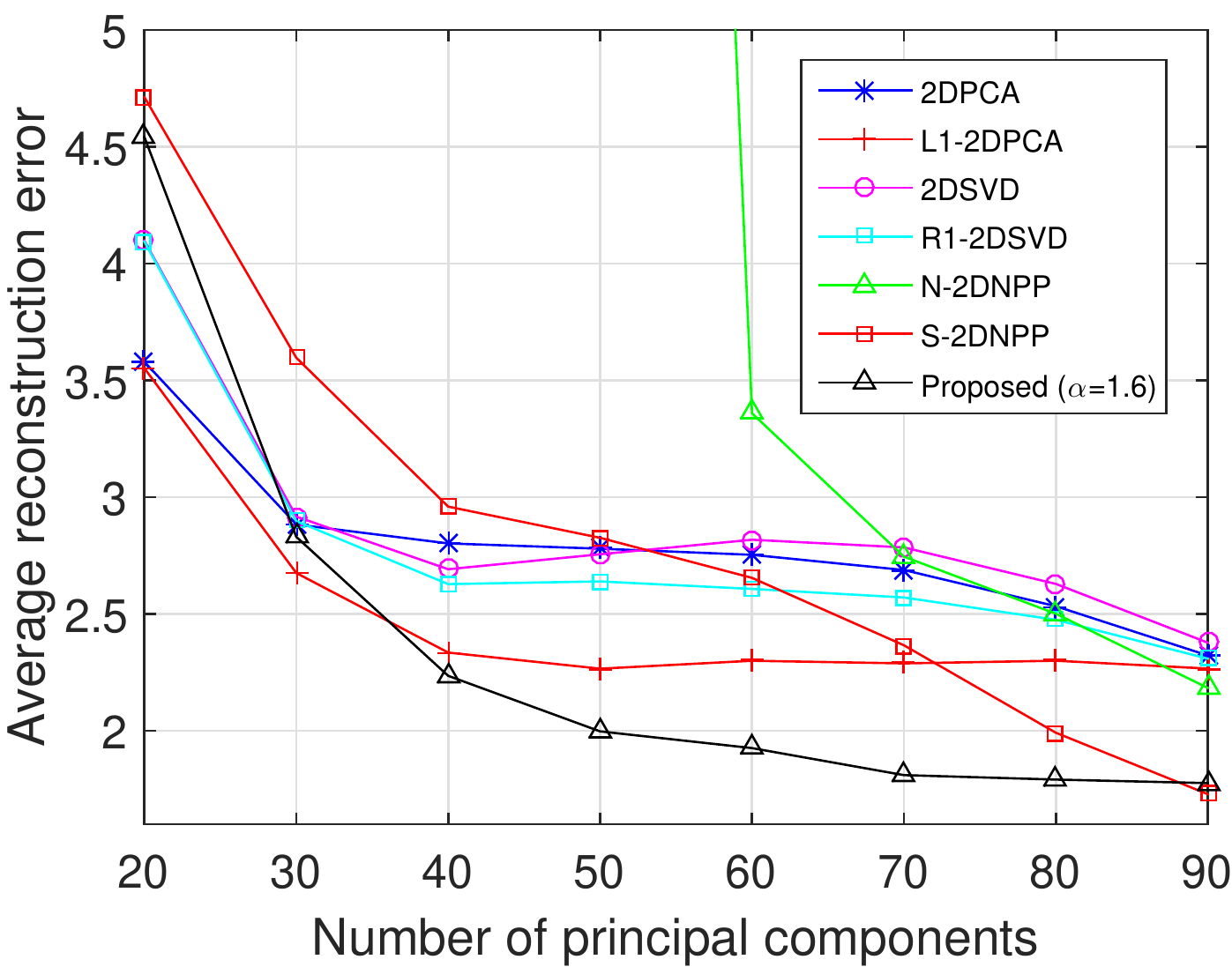}}
  \end{center}
   \caption{Average reconstruction errors of different algorithms on the Yale database. (a) The variation trend of reconstruction errors from the dataset with dummy images. (b) The variation trend of reconstruction errors from the dataset with block outliers.}
\end{figure*}
\vspace{-1cm}
\begin{figure*}
\begin{center}
\hspace{-0.3cm}
\vspace{-0.5cm}
   \subfigure[]{\includegraphics[width=0.51\columnwidth]{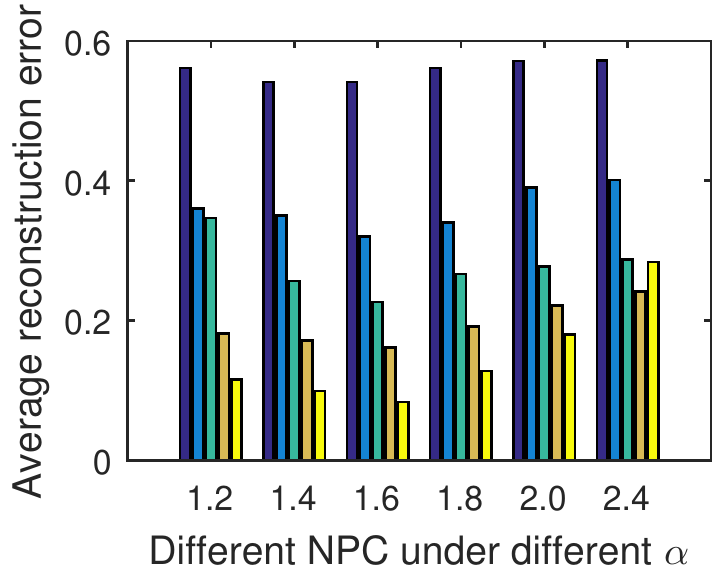}}
  \subfigure[]{\includegraphics[width=0.51\columnwidth]{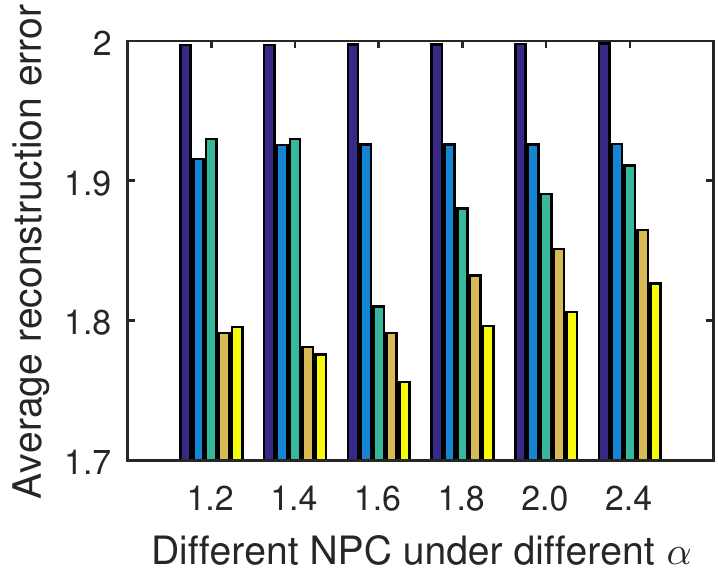}}
  \subfigure[]{\includegraphics[width=0.5\columnwidth]{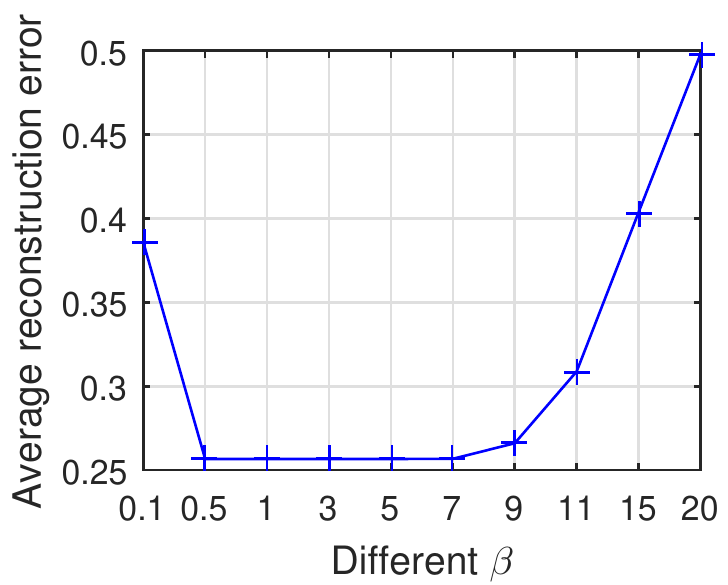}}
  \subfigure[]{\includegraphics[width=0.51\columnwidth]{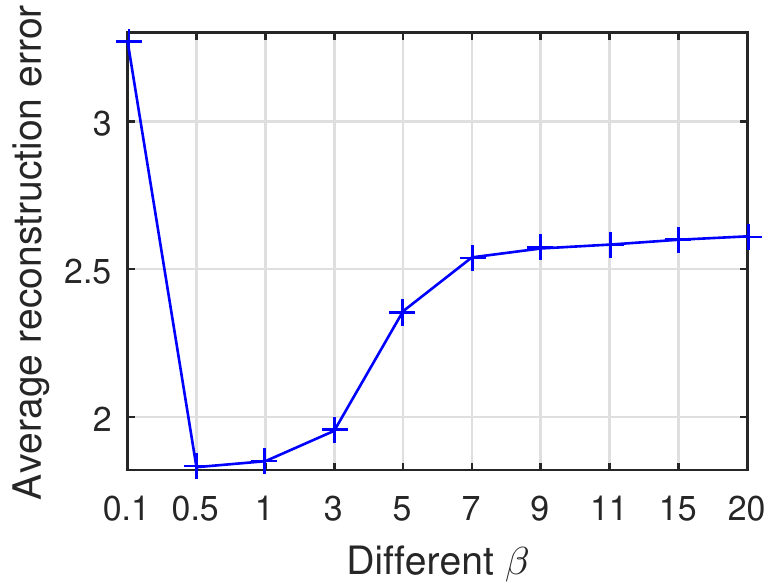}}
  \end{center}
   \caption{Average reconstruction errors of the proposed algorithm with different $\alpha$ and $\beta$ values on the Yale database. (a) reconstruction results from the dataset with dummy images under different NPC and $\alpha$ with $\beta=0.8$. (b) reconstruction results from the dataset with block outliers under different NPC and $\alpha$ with $\beta=0.8$. (c) reconstruction results from dataset with dummy images under different $\beta$ with $\alpha=1.6$ and $\text{NPC}=70$. (d) reconstruction results from the dataset with block outliers under different  $\beta$ with $\alpha=1.6$ and $\text{NPC}=70$. For (a) and (b), $\alpha$ varies from 1.2 to 2.4, and for each $\alpha$, the reconstruction error varies with the NPC varying from 50 to 90, corresponding to five colored bars.}
\end{figure*}
\begin{figure}
\begin{center}
\hspace{-0.3cm}
\vspace{-0.5cm}
   \subfigure[]{\includegraphics[width=0.5\columnwidth]{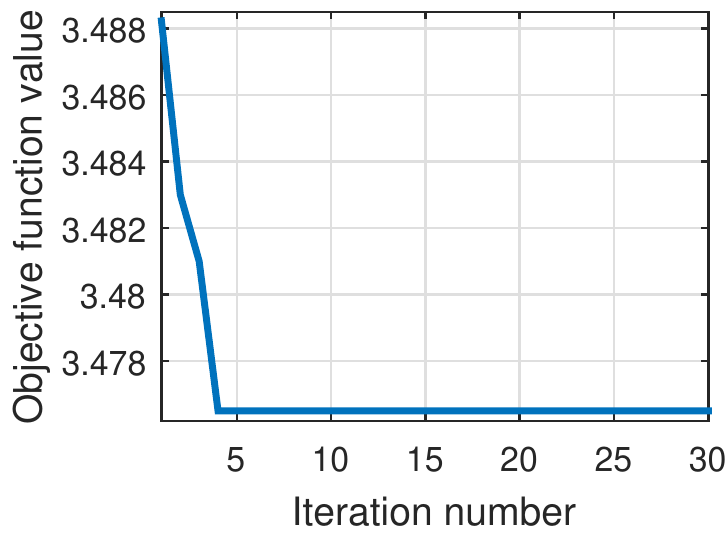}}
    \hspace{-0.5cm}
  \subfigure[]{\includegraphics[width=0.5\columnwidth]{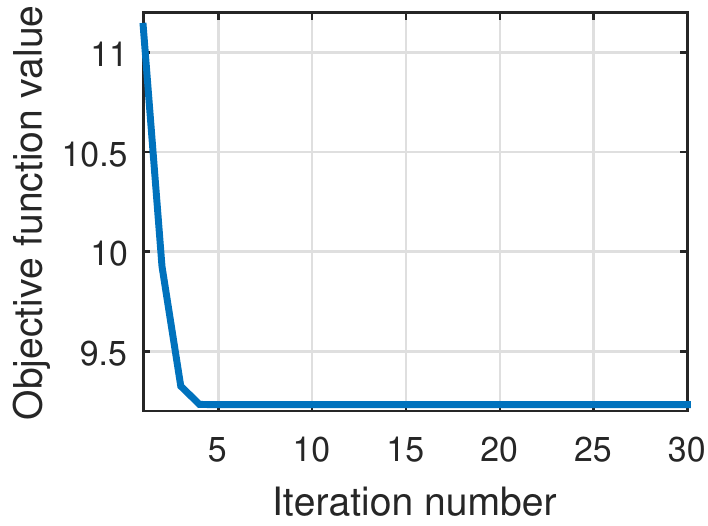}}
  \end{center}
   \caption{Convergence curves of the proposed method. (a) convergence curve from the dataset with dummy images. (b) convergence curve from the dataset with block outliers.}
\end{figure}
\subsection{Image Reconstruction}
In this part, we carry out experiments for image reconstruction on both data with sample outliers and block outliers. To demonstrate the convergence rate of the proposed algorithm, we plot the convergence curves of both cases in Fig. 4 which shows that the proposed method can converge within several iterations.
\subsubsection{Sample Outliers}
We randomly generate 30 dummy images (using random values between 0 and 1 as the outlier images and add them to the Yale dataset. So the number of inliers and outliers used in the training state is 165 and 30, respectively. A similar outlier generating strategy can also be found in~\cite{he2011robust}. When calculating the reconstruction errors, we exclude the calculation for outliers and just focus on reconstructing inlier images from the learned features. Fig. 2(a) shows the average reconstruction errors of the proposed algorithm and the benchmarks with the number of principal components (NPC) varies from 20 to 90. The curves for the reconstruction errors from all the methods decrease with the increase of the number of principal components, which shows that the reconstructed images gradually approach the original images when NPC increases. The proposed method achieves the lowest reconstruction error when NPC varies from 30 to 90 with $\alpha=1.6$ and $\beta=0.8$ because the proposed framework has the superiority to others in minimizing the outlier influence in the training data. These analyses show that the proposed method achieves the state-of-the-art performance and outperforms other methods.

To verify the effectiveness of the proposed method when $\alpha\neq2$, we also show the reconstruction error bar chart of our algorithm under different $\alpha$ with $\beta=0.8$ in Fig. 3(a). The x-axis shows the value of different $\alpha$ when the NPC (shown as five colored bars) varies from $50$ to $90$ under each $\alpha$. We can see from this figure that the performance of the proposed method with $\alpha<2$ outperforms that with $\alpha=2$, and the reconstruction error reaches its lowest level when $\alpha$ is 1.6. To verify the effect of $\beta$ on the reconstruction error, we plot the curve of reconstruction error under different $\beta$ with $\alpha=1.6$ and $\text{NPC}=70$ in Fig. 3(c). As one can see, the performance becomes worse when $\beta$ is too small or too large. In this experiment, the proposed algorithm reaches its best results when $\beta$ is in the range of $[0.5,7]$.
\subsubsection{Block Outliers}
We randomly select 30 facial images among the 165 images in the Yale database and partially block each of them by a rectangular area with random black and white dots~\cite{he2011robust}\cite{chen2017robust}. So the number of inliers and outliers used in this experiment are 135 and 30, respectively. Fig. 2(b) displays the average reconstruction error curves of different algorithms. When the number of principal components is small (less than 30), the average reconstruction errors of our method are higher than that of other algorithms. However, with the increase of the number of principal components, the reconstruction error of the proposed algorithm continuously decreases and finally reaches the lowest value among all the competing algorithms. This is because the proposed method has the superiority in weakening the influence from outlier images, and thus the learned projected matrices (feature) from our method contain less outlier information than from other methods. The proposed method gives a small weight if a sample is an outlier in the alternative optimization process. In the ideal case, the weights corresponding to the outliers would be zeros, which means that the outlier information will be removed from the training dataset. Thus the eigenvalues obtained from our method contain no information from outliers. Fig. 3(b) shows the average reconstruction error of the proposed algorithm under different NPC and $\alpha$ with $\beta=0.8$, which is consistent with the results in Fig. 3(a), that is the performance with $\alpha<2$ is better than that with $\alpha=2$, and the proposed algorithm reaches its best performance when $\alpha=1.6$. Fig. 3(d) gives the results of the proposed algorithm under different $\beta$ with $\text{NPC}=70$ and $\alpha=1.6$, which shows that the proposed algorithm reaches its best results when $\beta$ is in the range of $[0.5,1]$.
\subsubsection{Color Image Reconstruction}
To verify the effectiveness of the proposed algorithm in processing higher order tensor data, in this section, we test the proposed method on the color images of the LFW database with outliers. We select a total of 21 facial images under the name of ``Amelie Mauresmo" as inliers for training. Ten dummy images are created as outliers. Fig. 5 shows the average reconstruction error under different $\alpha$ and $\beta$ with NPC fixed at 40. These two figures show that the proposed algorithm with $\alpha = 3$ and $\beta \in [0.5, 11]$ yields the best performance, and the proposed algorithm is more sensitive to a changing $\alpha$ than to a changing $\beta$ for color image reconstruction. Fig. 6 displays the reconstruction ability of the proposed algorithm on two different images with NPC varying from 10 to 50. With the increase of NPC, the reconstruction quality increases, and it finally approaches that of the ground truth image in the 6th column, which further shows that the proposed algorithm is less sensitive to outliers.
\begin{figure}
\begin{center}
\hspace{-0.3cm}
\vspace{-0.5cm}
   \subfigure[]{\includegraphics[width=0.5\columnwidth]{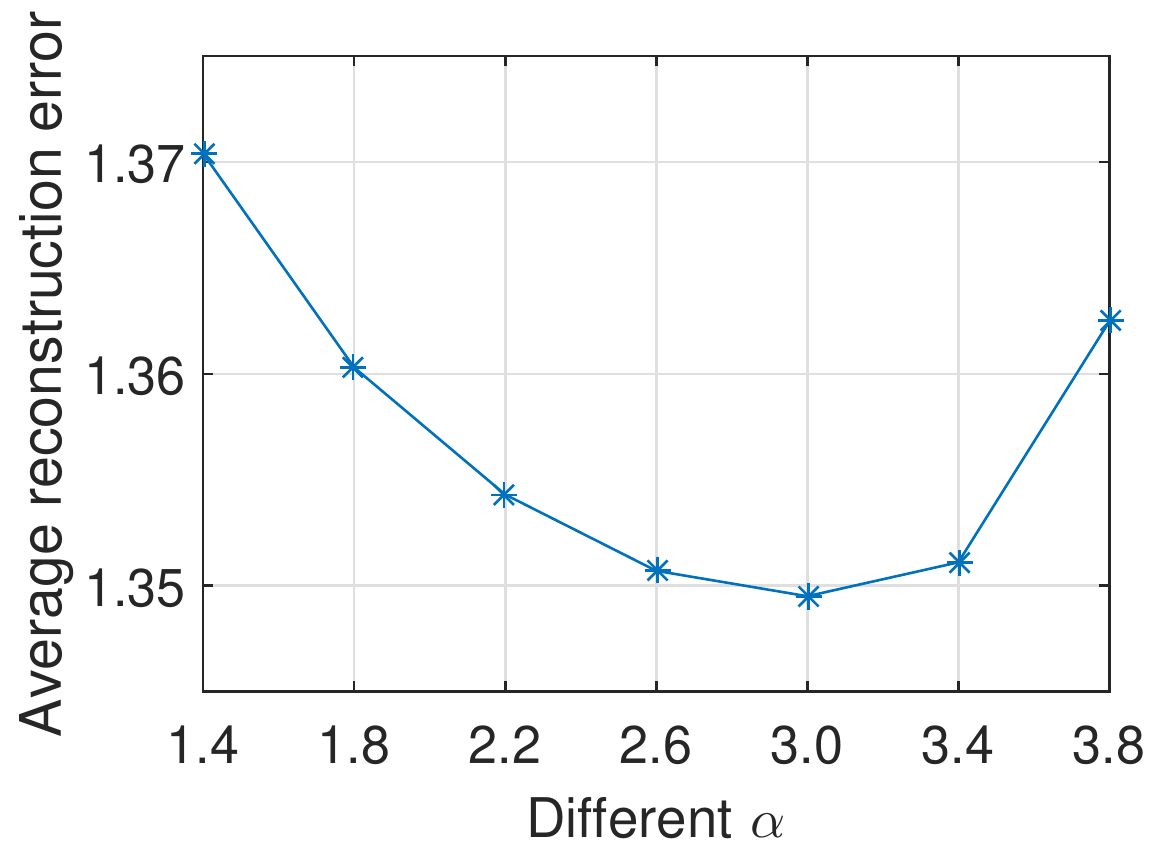}}
    \hspace{-0.3cm}
  \subfigure[]{\includegraphics[width=0.5\columnwidth]{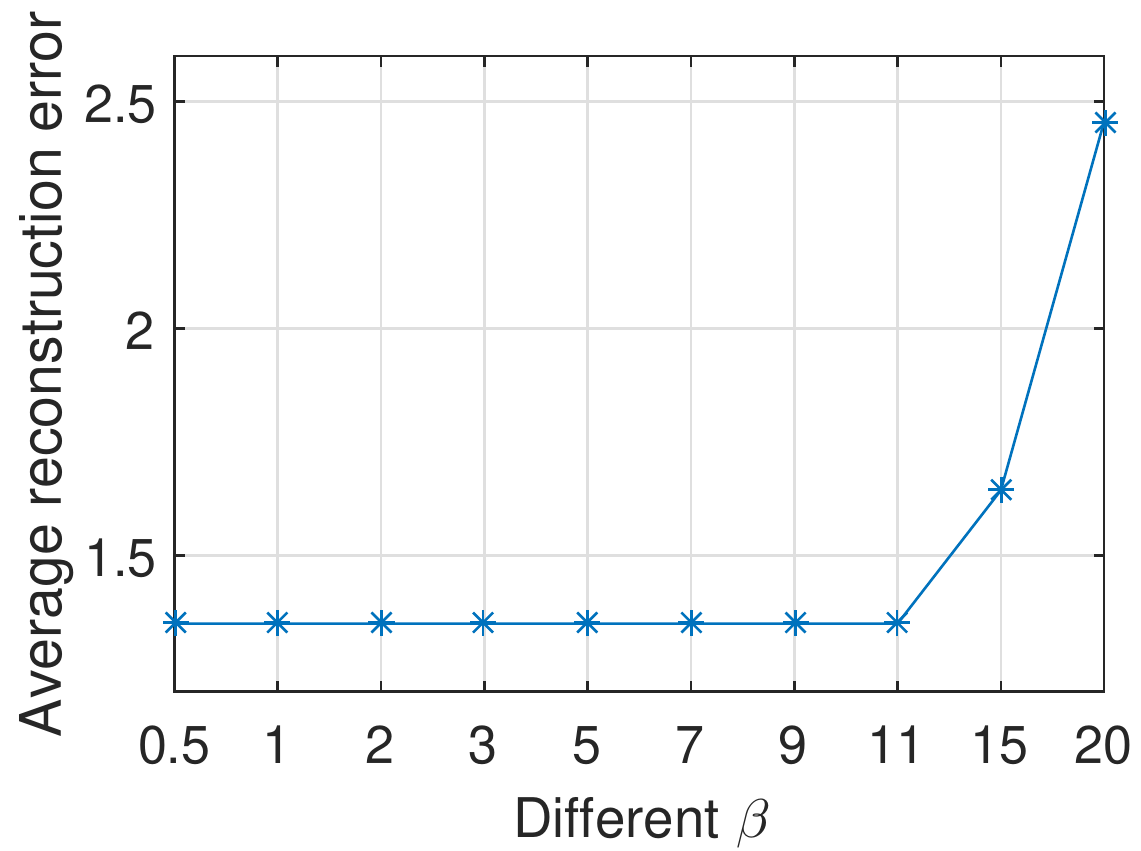}}
  \end{center}
   \caption{Average reconstruction error of the proposed method on LFW database. (a) Average reconstruction error of the proposed method under different $\alpha$ with $\text{NPC} = 40$ and $\beta = 0.8$. (b) Average reconstruction of the proposed method under different $\beta$ with $\text{NPC} = 40$ and $\alpha = 3$}
\end{figure}
\begin{figure}
\begin{center}
\hspace{-0.3cm}
\vspace{-0.5cm}
   \includegraphics[width=0.9\columnwidth]{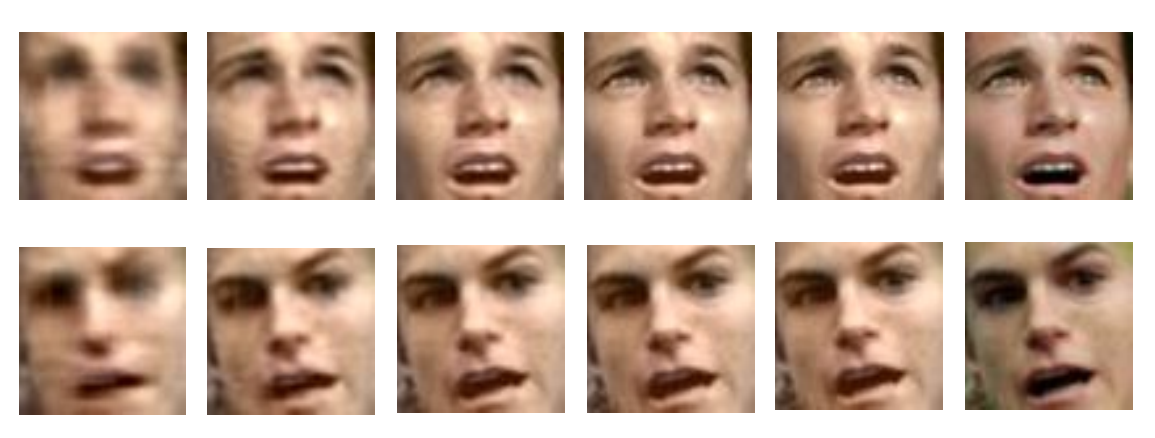}
  \end{center}
   \caption{Reconstructed images of the proposed method under different NPC with $\alpha = 3$ and $\beta = 0.8$. The 1st-5th column: reconstructed images with NPC varying from 10 to 50. The 6th column: ground truth.}
\end{figure}
\subsection{Image Classification}
\subsubsection{Handwritten Digit Recognition}
In this experiment, we evaluate the effectiveness of the proposed algorithm for image classification on the MNIST dataset. Normally, the outliers are typically far away from the normal data. Thus, we use the original images weighted by a magnitude to simulate outliers, i.e., ${X}^{\text{outlier}}=m{X}^{\text{org}}$, where ${X}^{\text{outlier}}$, ${X}^{\text{org}}$, and $m$ are the simulated outlier image, original image, and the magnitude of the outlier, respectively. Similar to~\cite{he2011robust}, we randomly select $5\%$ of the samples in the training sets as outliers, the remaining $95\%$ of the samples as inliers, and $m=50$ as the magnitude of outliers for all algorithms. To reduce the statistical deviations, all experimental results are reported over 20 random trials. Algorithm 2 in Section V is used for all the methods for image classification. Since other methods do not have the mechanism to distinguish outliers, we set the weights to $1$ when using Algorithm 2 for other benchmarks.

Table I compares the classification results from five different methods on four different sizes of datasets. For all the algorithms, the NPC is set to $15$, that is $k_1=k_2=15$. Numerical results in Table I show that the recognition accuracies of all the algorithms increase when the number of training samples increases. Compared with other methods, the classification rates of the proposed algorithm always achieves the best performance (marked in bold) under different NPC. To check the effect of different levels of representation error on the recognition accuracy, we also compare results of the proposed method under different $\alpha$ in Table I. The results show that the recognition accuracy of the proposed method is much better under the case of $\alpha>2$, and that the accuracy reaches the best with $77.83\%$ when $\alpha=4$.

To give more intuitive analysis of the performance of the proposed method under different parameters, we visually display the change of recognition rates with different $\alpha$ and $\beta$ in Fig. 7(a). The number of training samples and the percentage of outliers that are used in this experiment are set to $2000\times 10$ and $5\%$, respectively. With $\alpha=1$, the accuracy for the proposed method remains the lowest with each different $\beta$. The accuracy increases with the increase of $\alpha$ when fixing a $\beta$ and reaches the highest level with $\alpha=4$ and $\beta=0.8$. The accuracies increase when $\beta$ increases from $0.2$ to $0.8$ with a fixed $\alpha$ and slightly decrease when $\beta$ continues to increase.

\begin{figure}
\begin{center}
\hspace{-0.3cm}
\vspace{-0.5cm}
   \subfigure[]{\includegraphics[width=0.5\columnwidth]{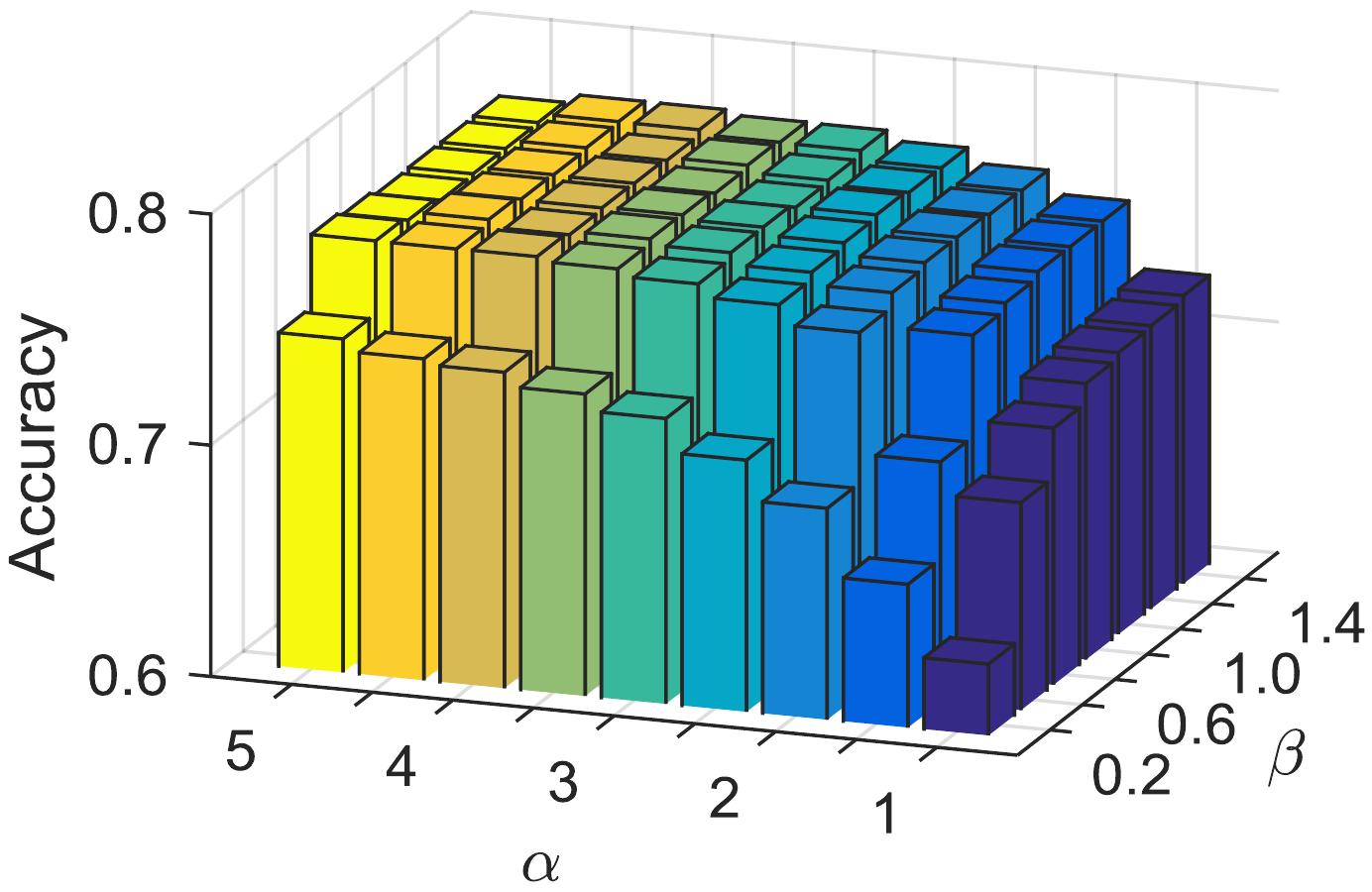}}
    \hspace{-0.2cm}
  \subfigure[]{\includegraphics[width=0.48\columnwidth]{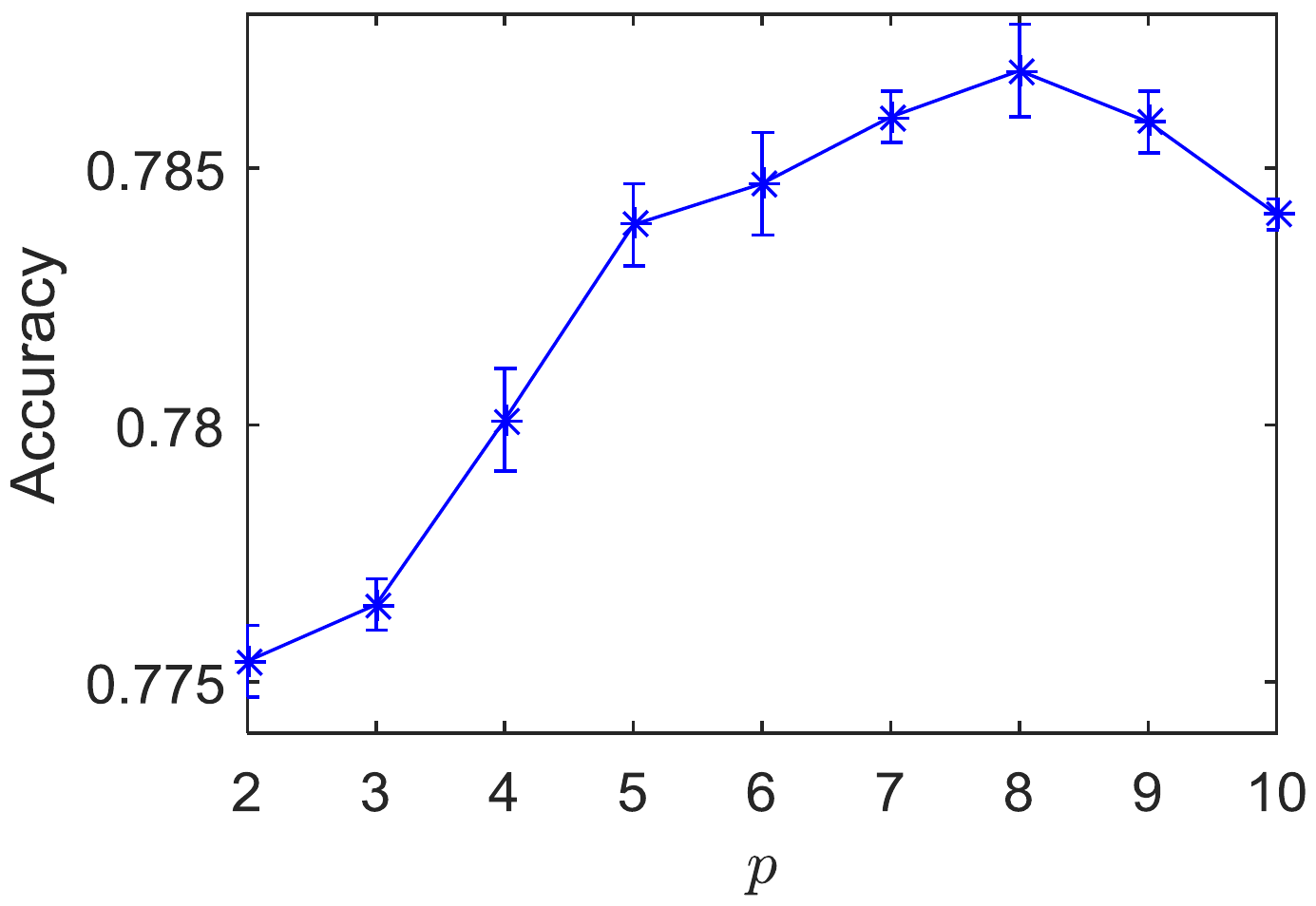}}
  \end{center}
   \caption{Classification accuracy of the proposed method with different parameters on the MNIST handwritten digit dataset. (a) Classification accuracy of the proposed method with different $\alpha$ and $\beta$. (b) Classification rate versus different $p$ values.}
\end{figure}
The features we used in the above recognition experiments for the proposed method is the result from a second order statistical cost function. To verify whether the performance of non-second order statistical objective function is effective as claimed in Section VI, we plot the classification rates with different $p$ values in Fig. 7(b). In this figure, the classification rate continuously increases with the increase of $p$ and achieves at $78.74\%$ with $p=10$, which shows that the proposed method with a $p$ value greater than $2$ performs better than that with a normally used value $p=2$.
\begin{figure}
\begin{center}
\hspace{-0.3cm}
\vspace{-0.5cm}
   \subfigure[]{\includegraphics[width=0.5\columnwidth]{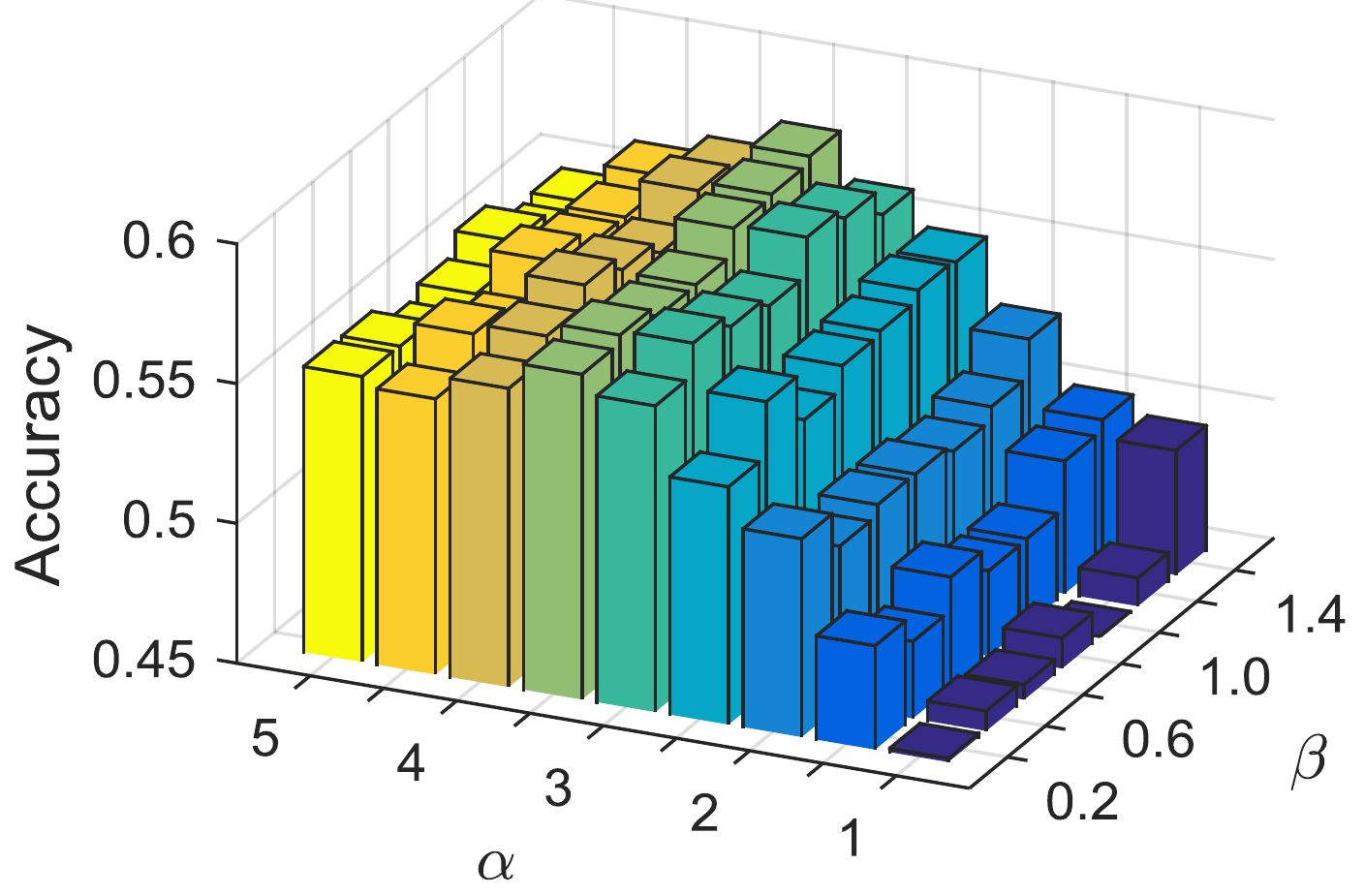}}
    \hspace{-0.2cm}
  \subfigure[]{\includegraphics[width=0.48\columnwidth]{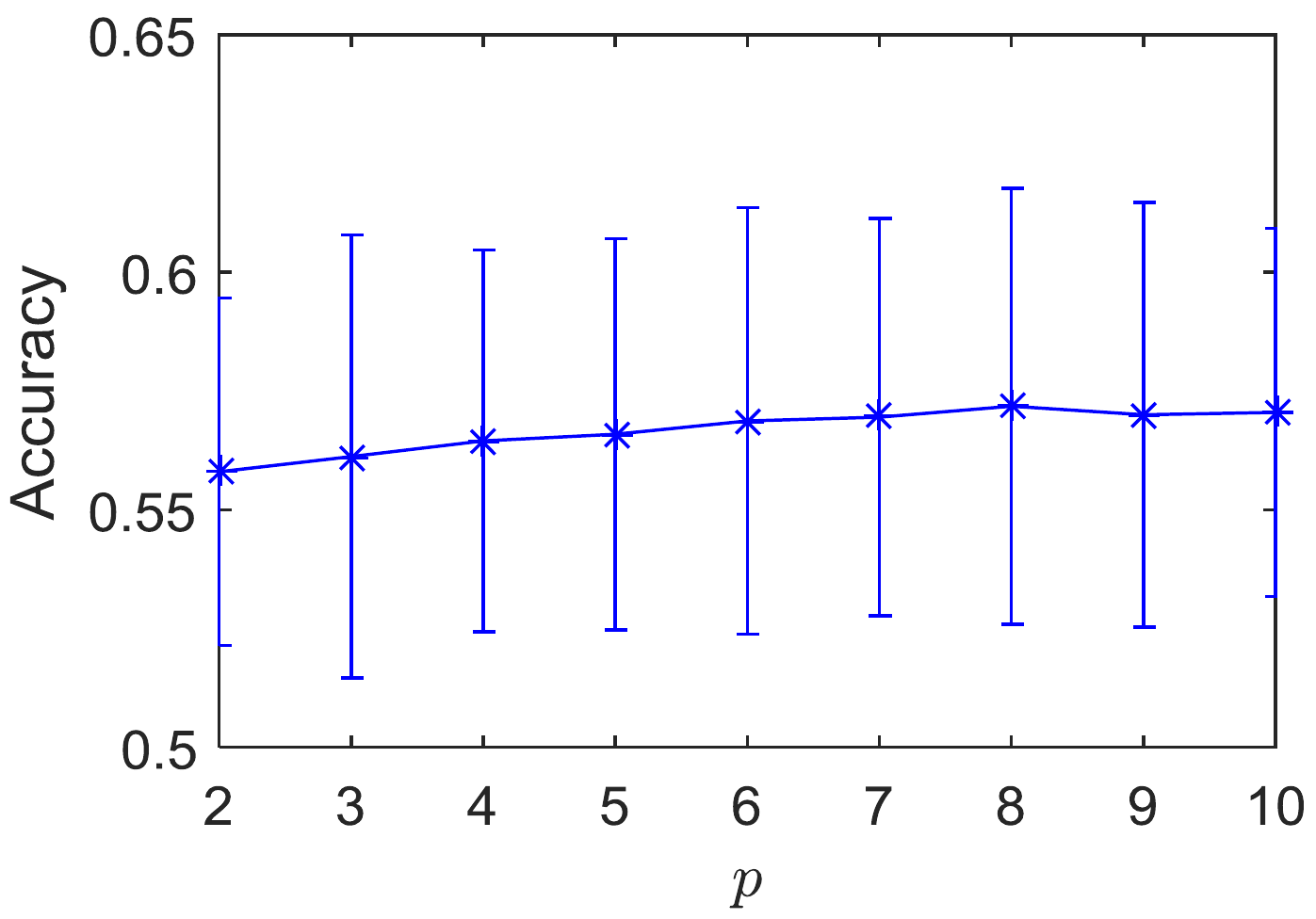}}
  \end{center}
   \caption{Classification accuracy of the proposed method with different parameters on the LFW database. (a) Classification accuracy of the proposed method with different $\alpha$ and $\beta$. (b) Classification rate versus different $p$ values.}
\end{figure}
\subsubsection{LFW Database Recognition} To examine the robustness of our method in an unconstrained environment, we evaluate its performance on the LFW database. First, we test all the algorithms on the gray scale images, and then our algorithm is used to classify RGB color images to verify the outlier resistance ability in higher order tensor space. The NPC is set to 40 for all experiments. We select the subjects that have more than 80 samples for this experiment. Five subjects are selected, and their names are ``Colin.Powell'', ``Donald.Rumsfeld'', ``George.W.Bush'', ``Gerhard.Schroeder'', and ``Tony.Blair'', and the number of samples for each subject are 236, 121, 530, 109, 144. We chose a different percentage of samples from each subject for training, and the remaining samples are used for testing. Ten dummy images are created as outliers and added to the training data. Table II gives the recognition accuracy of all the algorithms. To be consistent with Table I, we show the accuracy under the same parameters with Table I for the proposed algorithm but not the best accuracy of the proposed algorithm. To reduce the statistical deviations, all experimental results are reported over 20 random trials. We can see that, the proposed algorithm outperforms all the benchmarks. The proposed algorithm on color images achieves the best accuracy because the proposed method takes advantage of the higher order tensor decomposition in retaining the structure of color information from different channels. Fig. 8 analyzes the performance of the proposed algorithm under different parameters on the gray scale images. For the experiment in Fig. 8(a), the percentage of training samples is $70\%$. The accuracy gradually increases when $\alpha$ increases and reaches its best value at $57.74\%$ with $\alpha = 3.5$ and then the value gradually decreases. For Fig. 8(b), $\alpha=2.7$, $\beta=0.8$, and $\text{NPC}=40$ are applied. The accuracy increases from $55.80\%$ at $p=2$ to $57.18\%$ at $p = 8$.
 \begin{table*}  
\caption {The recognition accuracy of all the algorithms on the MNIST handwritten digit dataset with $5\%$ outliers: Average recognition accuracy (AC) $\pm$ standard derivation.}
 \centering
 \begin{tabular}{|p{100pt}<{\centering}| p{70pt}<{\centering} | p{70pt}<{\centering}| p{70pt}<{\centering} | p{70pt}<{\centering}|p{70pt}<{\centering}|}
   \hline
   \multirow{2}{*}{Methods}&
   \multicolumn{4}{c|}{Images per digit $\times$ $\sharp$ of digits}\\
   \cline{2-5}
  & $500\times10$ & $1000\times10$& $2000\times10$ & $3000\times10$\\
   \hline
   2DPCA & 0.3837 $\pm$ 0.1075 & 0.4287 $\pm$ 0.1027& 0.5296 $\pm$ 0.0901& 0.5845 $\pm$ 0.0938  \\
   \hline
   $L_1$-2DPCA & 0.3886 $\pm$ 0.0884 & 0.4520 $\pm$ 0.0927 & 0.5238 $\pm$ 0.1013 & 0.5594 $\pm$ 0.0866   \\
   \hline
   2DSVD  & 0.3745 $\pm$ 0.1168 & 0.4701 $\pm$ 0.0858 & 0.5411 $\pm$ 0.0922 & 0.5604 $\pm$ 0.0865  \\
   \hline
   $R_1$-2DSVD & 0.3556 $\pm$ 0.1146 & 0.4683 $\pm$ 0.1189 & 0.5481 $\pm$ 0.0799 & 0.5928 $\pm$ 0.0894  \\
   \hline
   N-2DNPP  &  0.4010 $\pm$ 0.0950  &  0.5012 $\pm$ 0.1128  &  0.5753 $\pm$ 0.1137  & 0.5773 $\pm$ 0.1149  \\
   \hline
   S-2DNPP &  0.3968 $\pm$ 0.0852  &  0.4927 $\pm$ 0.0739  &  0.5091 $\pm$ 0.0947  &  0.4906 $\pm$ 0.0821   \\
   \hline
  Proposed ($\alpha=1$, $\beta=0.8$) & 0.7138 $\pm$ 0.0023 & 0.7146 $\pm$ 0.0016 & 0.7163 $\pm$ 0.0009 & 0.7092 $\pm$ 0.0007  \\
   \hline
   Proposed ($\alpha=2$, $\beta=0.8$ ) & 0.7576 $\pm$ 0.0012 & 0.7596 $\pm$ 0.0012 & 0.7630 $\pm$ 0.0009 & 0.7614 $\pm$ 0.0006  \\
   \hline
   Proposed ($\alpha=3$, $\beta=0.8$ ) & 0.7626 $\pm$ 0.0015 & 0.7652 $\pm$ 0.0010 & 0.7724 $\pm$ 0.0007 & 0.7741 $\pm$ 0.0003  \\
   \hline
   Proposed ($\alpha=4$, $\beta=0.8$ ) & \textbf{0.7655 $\pm$ 0.0013} & \textbf{0.7702 $\pm$ 0.0008}& \textbf{0.7755 $\pm$ 0.0005} & \textbf{0.7783 $\pm$ 0.0005}  \\
   \hline
  Proposed ($\alpha=5$, $\beta=0.8$ )& 0.7564 $\pm$ 0.0019 & 0.7663 $\pm$ 0.0013 & 0.7723 $\pm$ 0.0010 & 0.7769 $\pm$ 0.0007 \\
   \hline
 \end{tabular}
 \end{table*}

 \begin{table*}  
\caption {The recognition accuracy of all the algorithms on the LFW database with 10 outliers: Average recognition accuracy (AC) $\pm$ standard derivation.}
 \centering
 \begin{tabular}{|p{115pt}<{\centering}| p{70pt}<{\centering} | p{70pt}<{\centering}| p{70pt}<{\centering} | p{70pt}<{\centering}|p{70pt}<{\centering}|}
   \hline
   \multirow{2}{*}{Methods}&
   \multicolumn{4}{c|}{Percentage of training samples}\\
   \cline{2-5}
  & $50\%$ & $60\%$& $70\%$ & $80\%$\\
   \hline
   2DPCA & 0.3857 $\pm$ 0.0726 & 0.4332 $\pm$ 0.0533 & 0.4704 $\pm$ 0.0494 & 0.4919 $\pm$ 0.0524  \\
   \hline
   $L_1$-2DPCA  & 0.4291 $\pm$ 0.0656 & 0.4703 $\pm$ 0.0519 & 0.4957 $\pm$ 0.0498 & 0.5046 $\pm$ 0.0382  \\
   \hline
   2DSVD  & 0.3129 $\pm$ 0.0734 & 0.3849 $\pm$ 0.0677 & 0.3964 $\pm$ 0.0499 & 0.4270 $\pm$ 0.0476 \\
   \hline
   $R_1$-2DSVD & 0.3195 $\pm$ 0.0932 & 0.3657 $\pm$ 0.0626 & 0.4012 $\pm$ 0.0459 & 0.4354 $\pm$ 0.0395  \\
   \hline
    N-2DNPP  &  0.4312 $\pm$ 0.0628  &  0.4779 $\pm$ 0.0587  &  0.5007 $\pm$ 0.0482 & 0.5133 $\pm$ 0.0364  \\
   \hline
   S-2DNPP &  0.4285 $\pm$ 0.0552  &  0.4849 $\pm$ 0.0424  &  0.4970 $\pm$ 0.0471  &  0.5176 $\pm$ 0.0375  \\
   \hline
  Proposed ($\alpha=2$, $\beta=0.8$) & 0.4865 $\pm$ 0.0478 & 0.5033 $\pm$ 0.0446 & 0.5080 $\pm$ 0.0393 & 0.5221 $\pm$ 0.0480  \\
   \hline
   Proposed ($\alpha=2.7$, $\beta=0.8$ ) & 0.5244 $\pm$ 0.0457 & 0.5406 $\pm$ 0.0415 & 0.5593 $\pm$ 0.0433 & 0.5717 $\pm$ 0.0470  \\
   \hline
   Proposed color ($\alpha=2$, $\beta=3.5$ ) & 0.4917 $\pm$ 0.0389 & 0.5220 $\pm$ 0.0380 & 0.5243 $\pm$ 0.0338 & 0.5383 $\pm$ 0.0370  \\
   \hline
  Proposed color ($\alpha=7$, $\beta=3.5$ )& \textbf{0.5472 $\pm$ 0.0397} & \textbf{0.5669 $\pm$ 0.0377} & \textbf{0.6107 $\pm$ 0.0441} & \textbf{0.6240 $\pm$ 0.0330} \\
   \hline
 \end{tabular}
 \end{table*}

\subsection{Image Clustering}
Theoretical analysis and experimental results~\cite{he2011robust}\cite{ding2004k}\cite{huang2008simultaneous} show that dimensionality reduction methods can be used as a preprocessing step to improve the accuracy (AC) of K-means clustering. In this experiment, we apply the proposed Corr-2DSVD algorithm and all the benchmark algorithms to a clustering problem on the ORL face database with outliers. The first 10 individuals are selected as the training dataset and thus 100 facial images in total are selected with 10 images per class. To simulate outliers, we randomly generate 20 outlier images and add them to the training set. The number of inlier and outlier are 100 and 20, respectively. After learning the dimension reduced features using all the competing algorithms, K-means clustering algorithm is used to evaluate the quality of features. If an algorithm does not give any special constraints to outliers, the learned features are more likely to be contaminated by outliers, thus the accuracy of clustering will be low. Certainly, the accuracy will be high if an algorithm has the ability to minimize the influence from outliers.

The 2DPCA based algorithms are one-sided transforms, and the 2DSVD based algorithms including our algorithm are two-sided transforms. Thus there is a little difference on the operations of both types of algorithms when using K-means. The details are given as follows:
\begin{figure*}
\begin{center}
\hspace{-0.3cm}
\vspace{-0.19cm}
   \subfigure[]{\includegraphics[width=0.52\columnwidth]{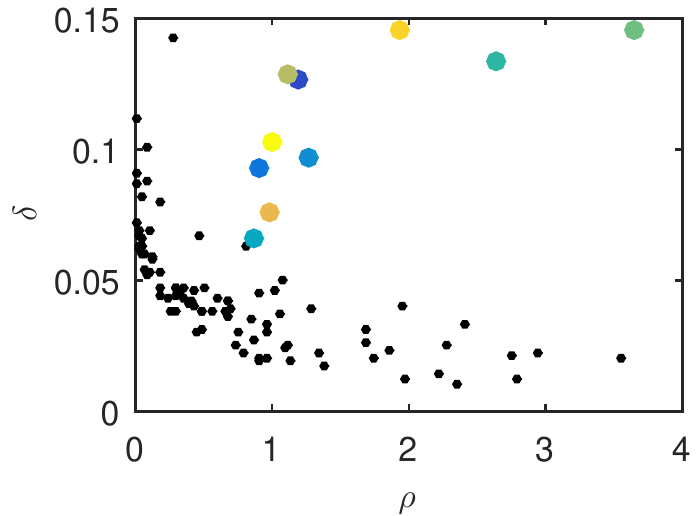}}
    \hspace{-0.3cm}
  \subfigure[]{\includegraphics[width=0.52\columnwidth]{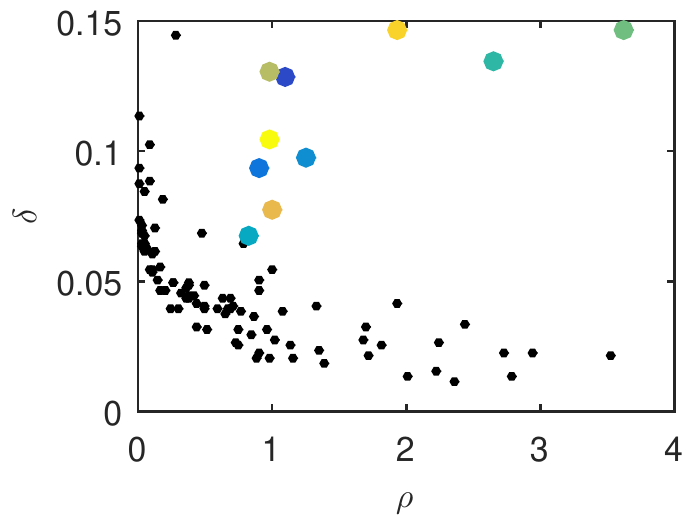}}
  \hspace{-0.3cm}
  \subfigure[]{\includegraphics[width=0.52\columnwidth]{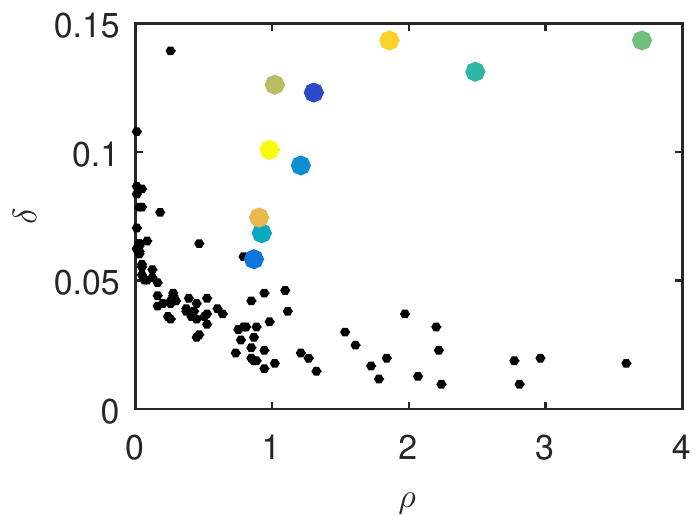}}
     \hspace{-0.3cm}
  \subfigure[]{\includegraphics[width=0.52\columnwidth]{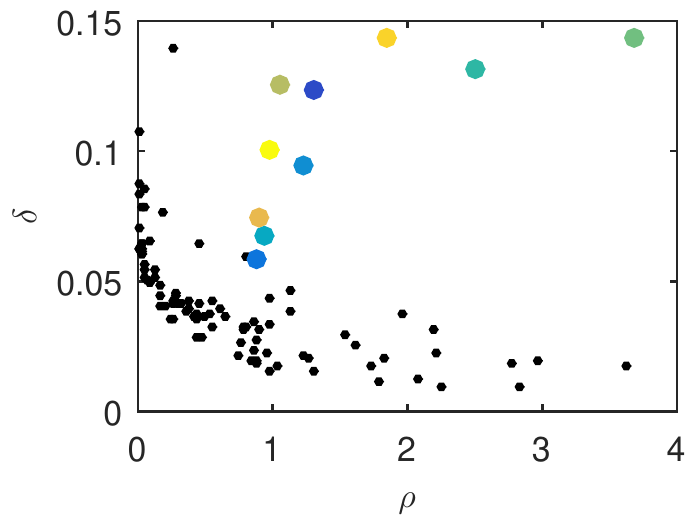}}
  \hspace{-0.3cm}
  \subfigure[]{\includegraphics[width=0.52\columnwidth]{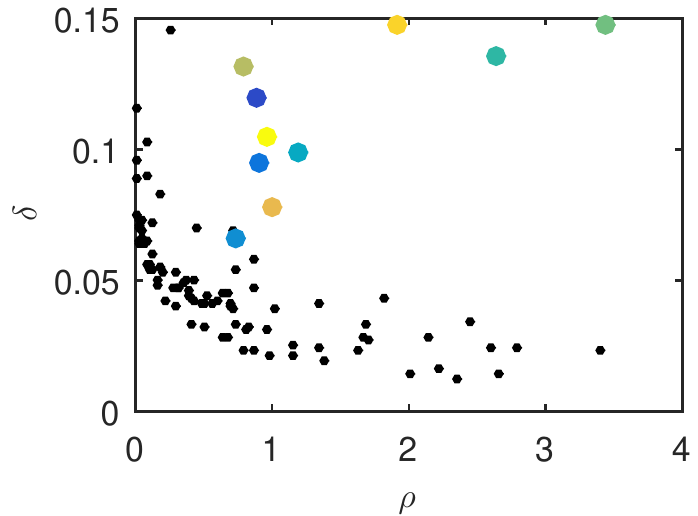}}
   \hspace{-0.3cm}
  \subfigure[]{\includegraphics[width=0.52\columnwidth]{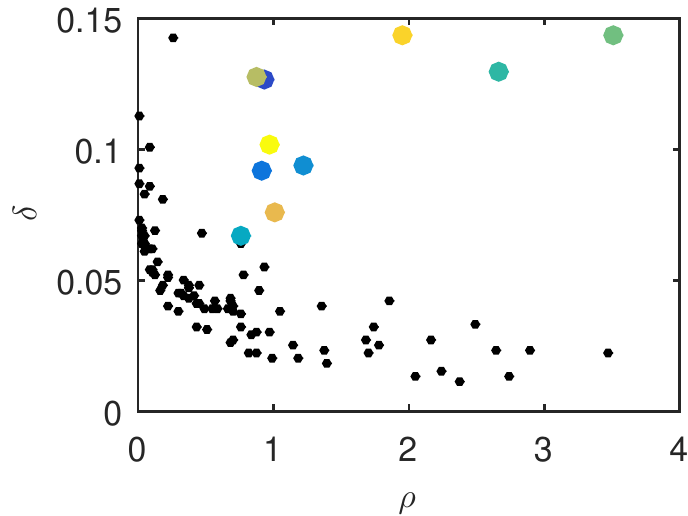}}
  \hspace{-0.3cm}
  \subfigure[]{\includegraphics[width=0.51\columnwidth]{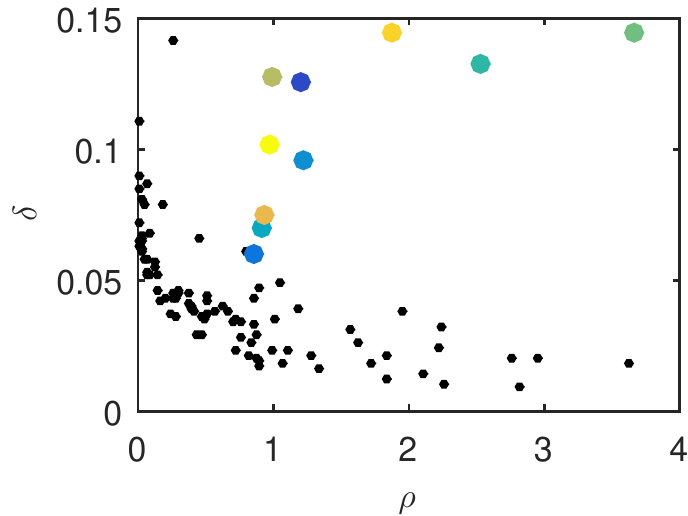}}
  \hspace{-0.32cm}
  \subfigure[]{\includegraphics[width=0.51\columnwidth]{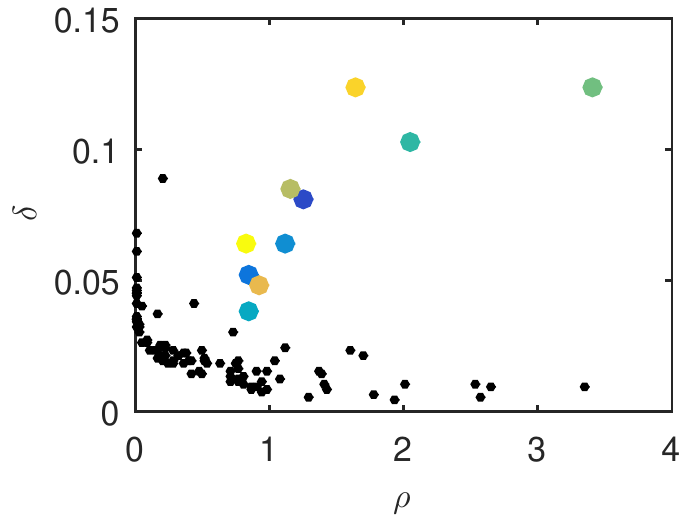}}
  \end{center}
   \caption{The decision graph for the first hundred images in the database obtained by different algorithms. The detected initial centers are colored. (a) 2DPCA, (b) $L_1$-2DPCA, (c) 2DSVD, (d) $R_1$-2DSVD, (e) N-2DNPP, (f) S-2DNPP, (g) proposed method ($\alpha=2$, $\beta=0.7$), (h) proposed method ($\alpha=6$, $\beta=0.7$).}
\end{figure*}
\begin{table*}
 \caption {K-means clustering results of subspaces learned from different algorithms on the first 100 faces of the ORL database: Average Clustering Accuracy (AC) $\pm$ Standard Deviation and Average normalized mutual information (NMI) $\pm$ Standard Deviation.}
 \centering
 \begin{tabular}{|p{90pt} <{\centering}| p{35pt}<{\centering} | p{70pt}<{\centering}| p{70pt}<{\centering} | p{70pt}<{\centering} |p{70pt}<{\centering}|}
    \hline
    \multicolumn{2}{|l|}{ \multirow{2}{*}{~~~~~~~~Methods and evaluation metrics}}&
   \multicolumn{4}{c|}{Number of principal components }\\
   \cline{3-6}
   \multicolumn{2}{|l|}{ } & $\text{NPC}=10$ & $\text{NPC}=30$ & $\text{NPC}=50$ & $\text{NPC}=70$ \\
   \hline
   \multirow{2}*{2DPCA}& AC & 0.8149 $\pm$ 0.0295 & 0.5955 $\pm$ 0.0405 & 0.7507 $\pm$ 0.0070 & 0.8157 $\pm$ 0.0167 \\
    & NMI & 0.8976 $\pm$ 0.0217 & 0.7598 $\pm$ 0.0246 & 0.8684 $\pm$ 0.0019 & 0.8864 $\pm$ 0.0046  \\
   \hline
   \multirow{2}*{$L_1$-2DPCA}& AC & 0.8219 $\pm$ 0.0241 & 0.6916 $\pm$ 0.0307 & 0.8240 $\pm$ 0.0226 & 0.8080 $\pm$ 0.0264 \\
    & NMI & \textbf{0.9027} $\pm$ \textbf{0.0177} & 0.8180 $\pm$ 0.0186 & 0.8875 $\pm$ 0.0159 & 0.8843 $\pm$ 0.0072  \\
   \hline
   \multirow{2}*{2DSVD} & AC & 0.7524 $\pm$ 0.0104  & 0.7451 $\pm$ 0.1203 & 0.7477 $\pm$ 0.0288& 0.7972 $\pm$ 0.0330 \\
    & NMI & 0.8644 $\pm$ 0.0072 & 0.8377 $\pm$ 0.0578 & 0.8557 $\pm$ 0.0188 & 0.8811 $\pm$ 0.0091 \\
    \hline
   \multirow{2}*{$R_1$-2DSVD}& AC &0.7531 $\pm$ 0.0087 & 0.7141 $\pm$ 0.1189 & 0.7538 $\pm$ 0.0264 & 0.8017 $\pm$ 0.0308\\
        & NMI & 0.8640 $\pm$ 0.0078 & 0.8223 $\pm$ 0.0586 & 0.8583 $\pm$ 0.0166 & 0.8825 $\pm$ 0.0085  \\
    \hline
   \multirow{2}*{N-2DNPP} & AC &
   0.8154 $\pm$ 0.0375 &  0.7944 $\pm$ 0.0865 & 0.7675 $\pm$ 0.0304&  0.7507 $\pm$ 0.0070  \\
    & NMI  &  0.8999 $\pm$ 0.0219 & 0.8791 $\pm$ 0.0294 & 0.8731 $\pm$ 0.0084 & 0.8684 $\pm$ 0.0019  \\
    \hline
   \multirow{2}*{S-2DNPP}& AC  &0.7417 $ \pm$ 0.0466 & 0.7303 $\pm$ 0.0294 & 0.7375 $\pm$ 0.0233 & 0.8178 $\pm$ 0.0159 \\
        & NMI  &  0.8359 $\pm$ 0.0311 & 0.8139 $\pm$ 0.0174 & 0.8364 $\pm$ 0.0116 & 0.8862 $\pm$ 0.0071 \\
   \hline
   \multirow{2}*{Proposed ($\alpha=2,\beta=0.7$)}& AC &0.7513 $\pm$ 0.0033 & 0.6868 $\pm$ 0.0377 & 0.7546 $\pm$ 0.0084 & 0.8101 $\pm$ 0.0244 \\
        & NMI & 0.8662 $\pm$ 0.0028 & 0.8151 $\pm$ 0.0229 & 0.8676 $\pm$ 0.0011 & 0.8848 $\pm$ 0.0067  \\
   \hline
   \multirow{2}*{Proposed ($\alpha=4,\beta=0.7$)}& AC &0.7615 $\pm$ 0.0257 & 0.7983 $\pm$ 0.0458 & 0.8348 $\pm$ 0.0615 & 0.8332 $\pm$ 0.0305 \\
        & NMI & 0.8662 $\pm$ 0.0123  & 0.8777 $\pm$ 0.0234 & 0.8890 $\pm$ 0.0255 & 0.8830 $\pm$  0.0129 \\
   \hline
   \multirow{2}*{Proposed ($\alpha=6,\beta=0.7$)}& AC &0.8375 $\pm$ 0.0651 & 0.8762 $\pm$ 0.0614 & \textbf{0.9319} $\pm$ \textbf{0.0442} &0.8535 $\pm$ 0.0661 \\
        & NMI & 0.8738 $\pm$ 0.0307 & 0.8975 $\pm$ 0.0298 & \textbf{0.9248} $\pm$ \textbf{0.0217} & 0.8910 $\pm$ 0.0302  \\
   \hline
   \multirow{2}*{Proposed ($\alpha=7,\beta=0.7$)}& AC &\textbf{0.8476} $\pm$ \textbf{0.0500} & \textbf{0.9027} $\pm$ \textbf{0.0546} & 0.9275 $\pm$ 0.0452 &\textbf{0.8607} $\pm$ \textbf{0.0569} \\
        & NMI & 0.8750 $\pm$ 0.0240 & \textbf{0.9104} $\pm$ \textbf{0.0268} & 0.9239 $\pm$ 0.0225 & \textbf{0.8911} $\pm$ \textbf{0.0287}  \\
   \hline
 \end{tabular}
 \end{table*}
\subsubsection{2DPCA and 2DNPP-based algorithm+K-means clustering}
2DPCA and 2DNPP based algorithms are applied to the $a\times b\times N$ image tensor for data compression with reduced dimension $a\times k$ for each sample, where $a\times b$ is the size of each image, $N$ is the number of training samples, and $k$ is the number of selected principal components. With the learned feature $W$, each projected sample can be described as $X_i^{\text{new}}=X_i \times W$. Then the K-means clustering method is used to cluster $X_i^{\text{new}}$.
\subsubsection{2DSVD-based algorithm+K-means clustering}
2DSVD is also applied to the same image tensor as that in the above 2DPCA case for data compression with reduced dimensions $k_1\times k_2$. In our experiment, the $k$, $k_1$, and $k_2$ are set to $50$. Then the K-means clustering method is used to cluster $M_l$. The calculation of $M_l$ can be found in Section IV.

The clustering performance of the traditional K-means algorithm is highly affected by the initial cluster center points. To minimize the influence to the cluster centers from outliers, in this experiment, we use the density searching based method~\cite{rodriguez2014clustering} as a preprocessing step to obtain the initial cluster centers. The algorithm in~\cite{rodriguez2014clustering} assumes that the cluster centers are surrounded by neighbors with a lower local density and that they are at a relatively large distance from any points with a higher local density, which guarantees that the clusters of different classes are far away from each other and that the data with the highest density can be selected as the initial cluster center. For each data point $i$, we need to compute two quantities: its local density $\rho_i$ and the distance $\delta_i$ which is measured by the minimum distance between point $i$ and any other point with a  higher density, i.e., $\delta_i=\underset{j:\rho_j>\rho_i}{\min}(d_{i,j})$, where $d_{i,j}$ is the distance between points $i$ and $j$. For the point with the highest density, the distance $\delta_i$ is set as the distance between the current sample and the sample with the largest distance to the current sample. Since $\delta_i$ is much larger than the typical nearest neighbor distance for points that are local or global maxima in the density, the clusters are recognized as points for which the value of $\delta_i$ is anomalously large. In Fig. 9, a density-versus-distance map, also known as a decision graph, is plotted for initial cluster center selection. In this figure, we find that the point with larger $\rho$ and $\delta$ values can be recognized as cluster centers (coloured points). The decision graphs of the proposed method in Fig. 9(e) and Fig. 9(f) with different parameters show better separation between the points with high and low densities because the proposed method has the superiority of outliers rejection, and thus its learned features have little influence from outliers.\\
\indent Since the learned features for the proposed algorithm are much cleaner than that of other benchmarks in the presence of outliers, the selected initial cluster centers from the proposed method are much more closer to the optimal centers than that from other methods. With the advantages in selecting initial clustering centers, the clustering accuracies of our method shown in Table III are apparently higher than that of other algorithms. All the results are reported over 100 random trials to reduce deviations. Two evaluation metrics AC and NMI introduced in Section V are used to quantitatively evaluate the performance of all the algorithms. The results in Table III show that the proposed method almost achieved the best results (marked in bold) in terms of both AC and NMI under different numbers of principal components, and that the proposed algorithm obtains better results with $\alpha>2$ and reaches the best results at $0.9319 \pm 0.0442$ for AC and $0.9248 \pm 0.0217$ for NMI with $\alpha=6$.
\begin{figure*}
\begin{center}
\hspace{0cm}
\vspace{0cm}
   \subfigure[]{\includegraphics[width=0.49\columnwidth]{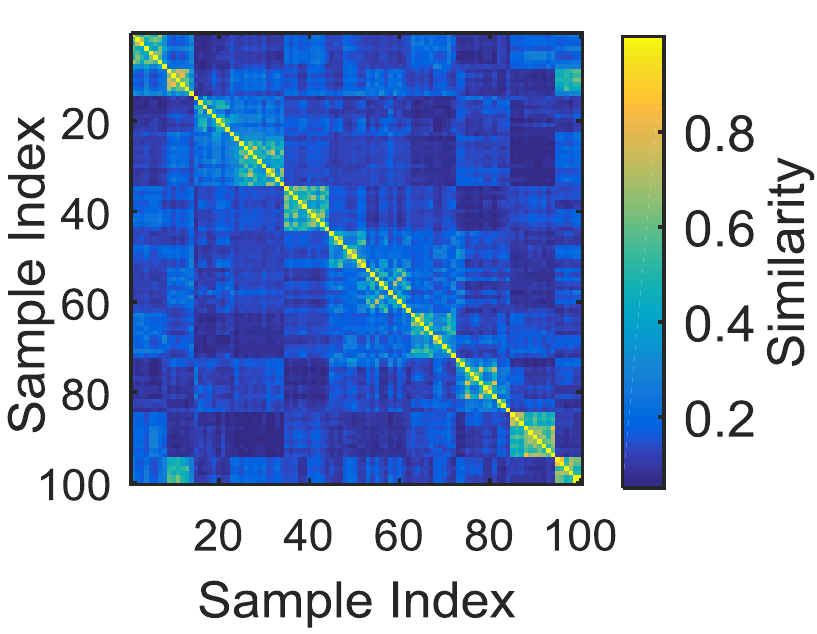}}
    \hspace{0cm}
  \subfigure[]{\includegraphics[width=0.49\columnwidth]{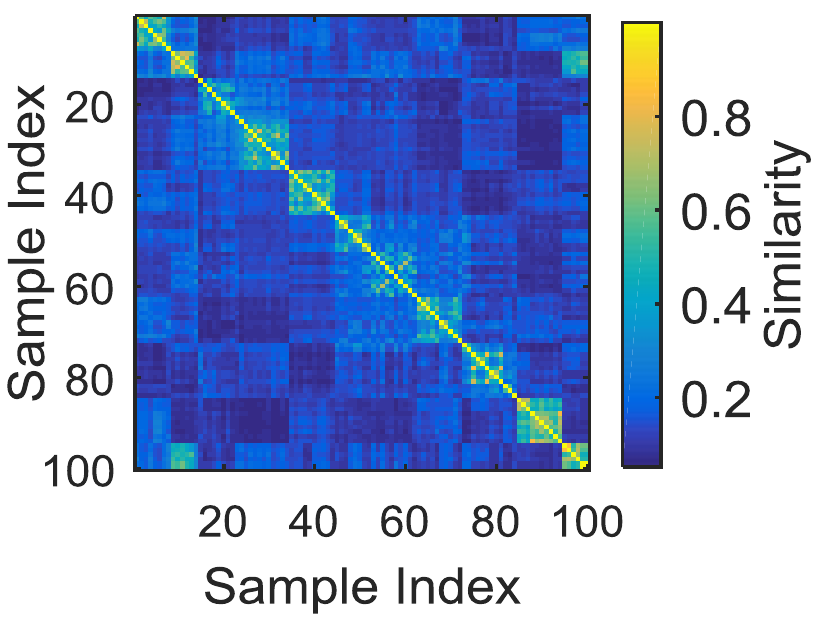}}
  \hspace{0cm}
  \subfigure[]{\includegraphics[width=0.49\columnwidth]{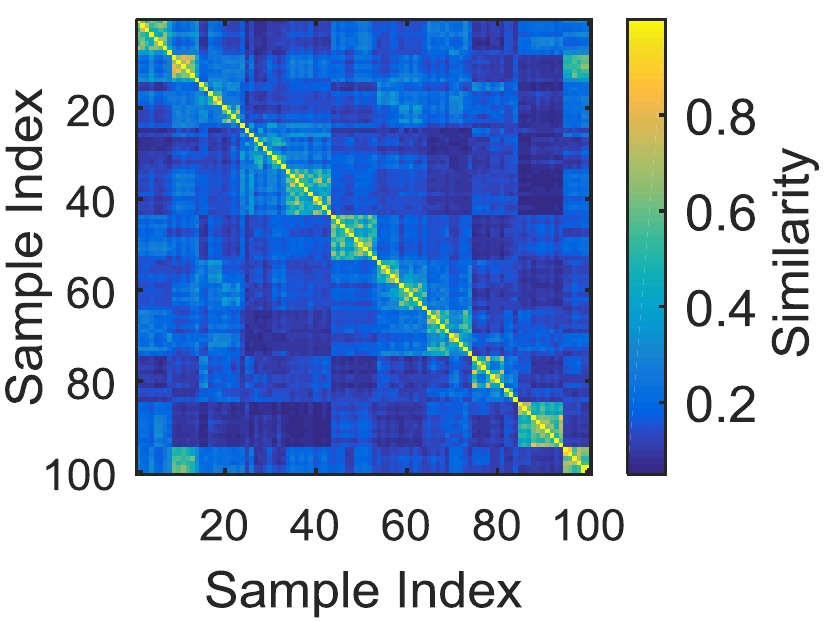}}
     \hspace{0cm}
  \subfigure[]{\includegraphics[width=0.49\columnwidth]{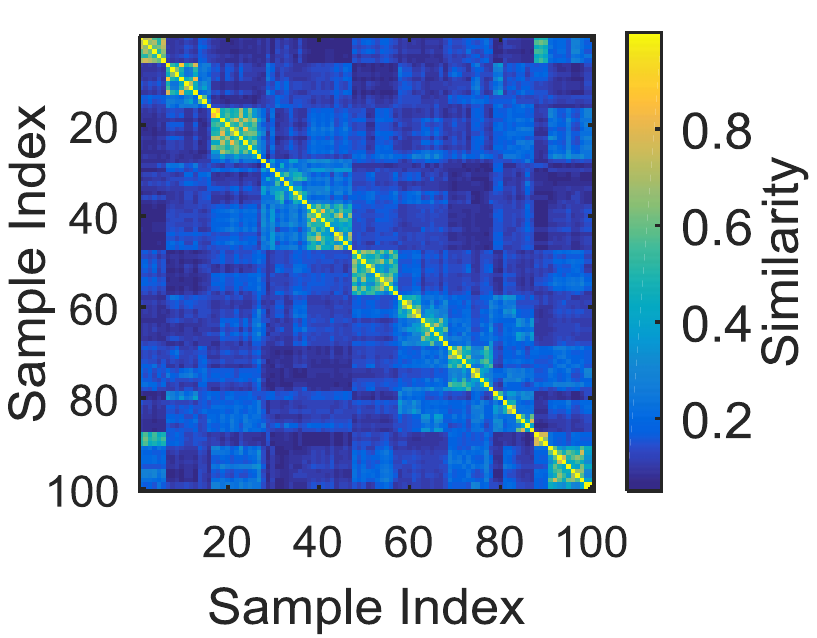}}
  \hspace{0cm}
  \subfigure[]{\includegraphics[width=0.49\columnwidth]{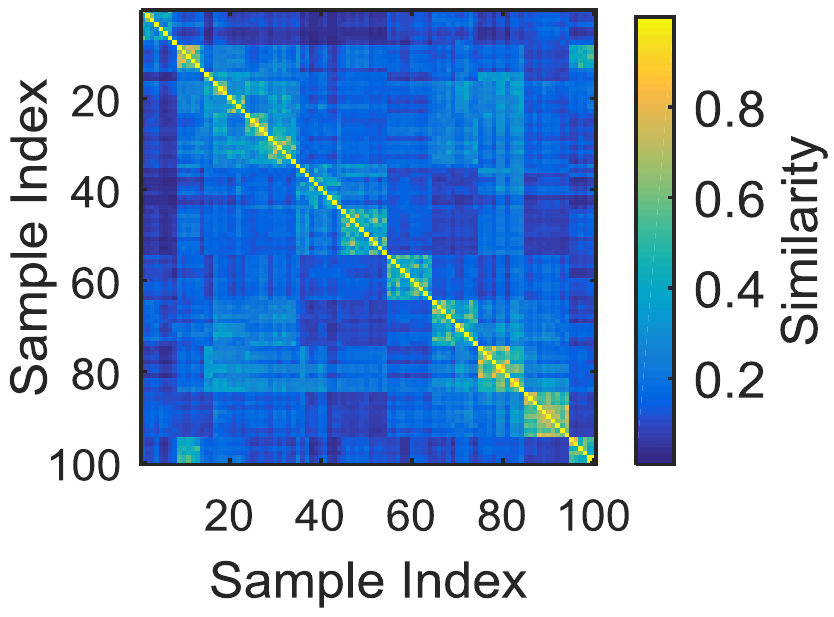}}
     \hspace{0cm}
  \subfigure[]{\includegraphics[width=0.49\columnwidth]{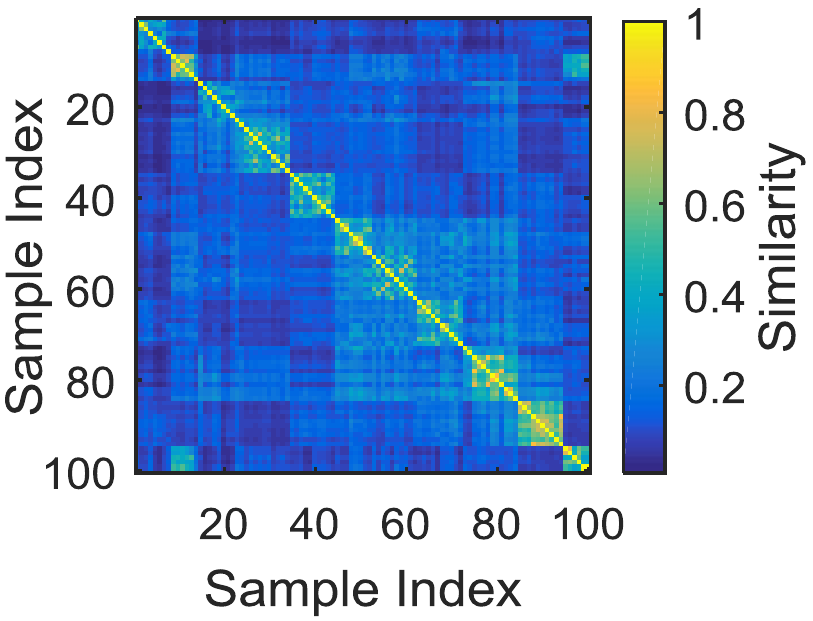}}
  \hspace{0cm}
  \subfigure[]{\includegraphics[width=0.49\columnwidth]{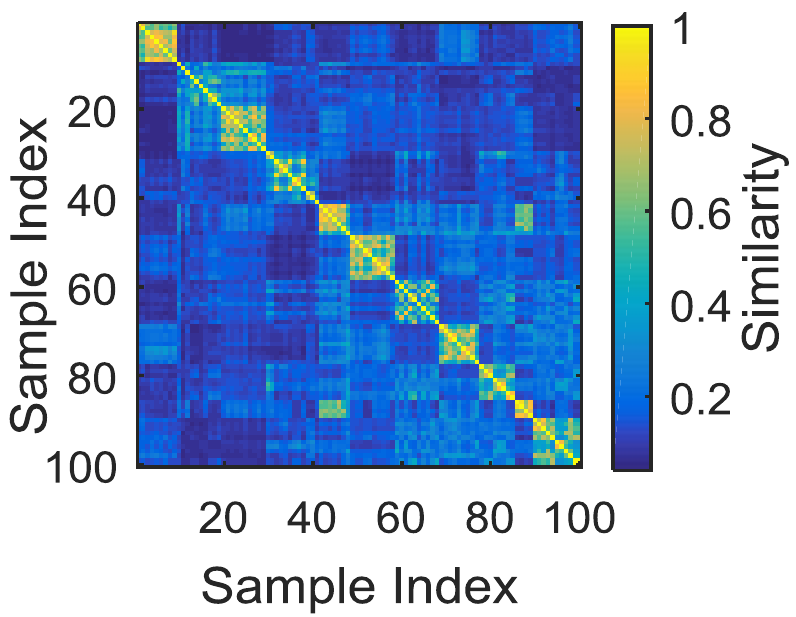}}
  \hspace{0cm}
  \subfigure[]{\includegraphics[width=0.49\columnwidth]{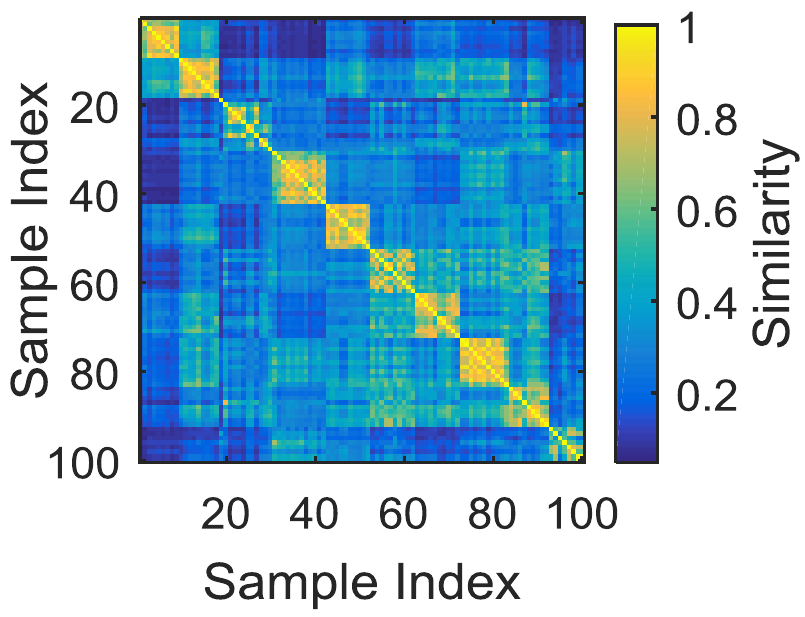}}
  \end{center}
  \vspace{-0.5cm}
   \caption{Visualization of the similarity matrices for different algorithms. (a) 2DPCA, (b) $L_1$-2DPCA, (c) 2DSVD, (d) $R_1$-2DSVD, (e) N-2DNPP, (f) S-2DNPP, (g) proposed method ($\alpha=2$, $\beta=0.7$), (h) proposed method ($\alpha=6$, $\beta=0.7$).}
\end{figure*}

To better explore the clustering results for each algorithm, we calculate and display the similarity matrix for each algorithm in Fig. 10. We reorder the similarity matrices with respect to cluster labels and inspect them visually. The light yellow square on the diagonal denotes the similarity level and clustering quality of the algorithms. The results in Fig. 10(h) from the proposed method with $\alpha=6$ show that our algorithm can correctly cluster most samples. There are only a few misclusterings in clustering classes $3$ and $9$. Our method with $\alpha=2$ also performs well only with minor clustering errors in classes $4$, $7$, and $8$. Based on above analyses, we can conclude that the proposed method achieves the state-of-the-art performance in clustering data when there are outliers, and that the performances of the proposed method with $\alpha>2$ are clearly better than that with $\alpha=2$.
\begin{figure}
\begin{center}
\hspace{-0.4cm}
\vspace{-0.5cm}
   \subfigure[]{\includegraphics[width=0.5\columnwidth]{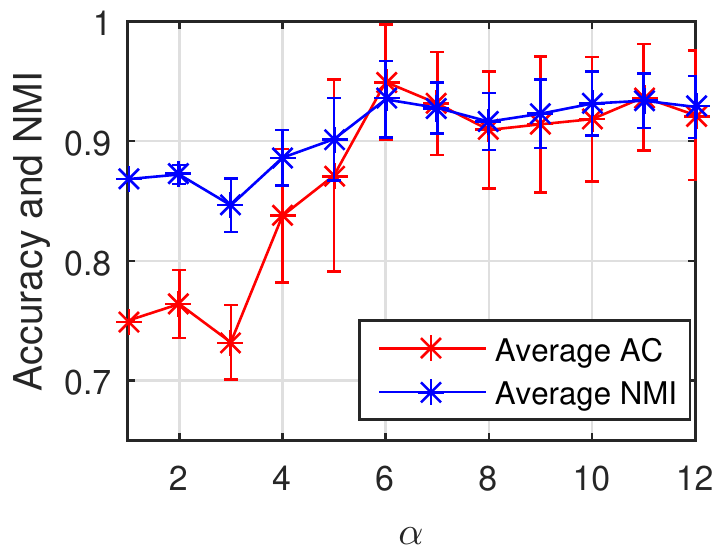}}
    \hspace{-0.2cm}
  \subfigure[]{\includegraphics[width=0.5\columnwidth]{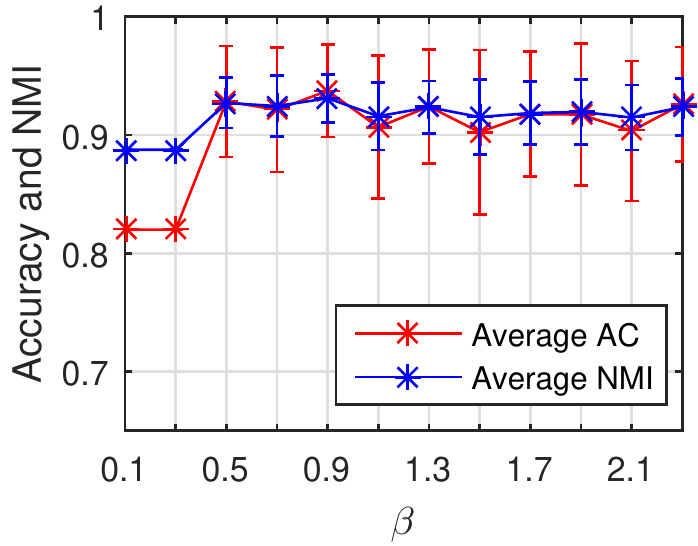}}
  \end{center}
   \caption{Average clustering accuracy and NMI under different $\alpha$ and $\beta$. (a) Average clustering accuracy and NMI under different $\alpha$ and $\beta=0.7$. (b) Average clustering accuracy and NMI under different $\beta$ with $\alpha=6$.}
\end{figure}

To analyze the effect of $\alpha$ and $\beta$ on AC and NMI, we plot the curves for AC and NMI in Fig. 11. Fig. 11(a) shows the average AC and NMI of the proposed algorithm under different $\alpha$ with $\beta$ fixed at $0.7$. In this figure, $\alpha$ varies from 1 to 12, and the curves for AC and NMI show a little fluctuation at $\alpha\in [1,4]$, then the values increase and reach their highest levels at $0.9319$ (AC) and $0.9248$ (NMI) at $\alpha=6$, after that there are no significant changes with the increase of $\alpha$. But when $\alpha$ is very large, e.g., greater than 15, the weight ($\omega$) for some samples will be $-\infty$, making it impossible for eigenvalue decomposition. In Fig. 11(b), we plot the curves for AC and NMI under different $\beta$ with $\alpha$ fixed at $6$, the values remain stable with only a little fluctuation when $\beta$ is greater than $0.5$. When $\beta$ is very large, the weight ($\omega$) for some samples will be $-\infty$, causing problems for eigenvalue decomposition.
\section{Conclusion}
This paper presented a new framework based on the Information Theoretic Learning to improve the robustness of tensor decomposition. By introducing the generalized correntropy to the traditional 2DSVD method, more flexible constraints are imposed on the representation error term, resulting in greater outlier resistance ability and improved performance in different image processing applications. Moreover, the data mean which is a key factor in experiments is updated automatically during iterations so that the optimized mean data will have less influence from outliers. A nearest center based classifier using the generalized correntropy is developed to further improve the classification performance. Experimental results on image reconstruction, image classification, and image clustering show that the proposed method has achieved the state-of-the-art performance and can be used as a robust approach for tensor decomposition.


\bibliographystyle{IEEEtran}

\begin{thebibliography}{10}
\providecommand{\url}[1]{#1}
\csname url@samestyle\endcsname
\providecommand{\newblock}{\relax}
\providecommand{\bibinfo}[2]{#2}
\providecommand{\BIBentrySTDinterwordspacing}{\spaceskip=0pt\relax}
\providecommand{\BIBentryALTinterwordstretchfactor}{4}
\providecommand{\BIBentryALTinterwordspacing}{\spaceskip=\fontdimen2\font plus
\BIBentryALTinterwordstretchfactor\fontdimen3\font minus
  \fontdimen4\font\relax}
\providecommand{\BIBforeignlanguage}[2]{{%
\expandafter\ifx\csname l@#1\endcsname\relax
\typeout{** WARNING: IEEEtran.bst: No hyphenation pattern has been}%
\typeout{** loaded for the language `#1'. Using the pattern for}%
\typeout{** the default language instead.}%
\else
\language=\csname l@#1\endcsname
\fi
#2}}
\providecommand{\BIBdecl}{\relax}
\BIBdecl

\bibitem{kwak2008principal}
N.~Kwak, ``Principal component analysis based on ${L}_1$-norm maximization,''
  \emph{IEEE Trans. Pattern Anal. Mach. Intell.}, vol.~30, no.~9, pp.
  1672--1680, 2008.

\bibitem{ding2006r}
C.~Ding, D.~Zhou, X.~He, and H.~Zha, ``${R}_1$-{PCA}: rotational invariant
  ${L}_1$-norm principal component analysis for robust subspace
  factorization,'' in \emph{Proc. Int. Conf. Mach. Learning}, 2006, pp.
  281--288.

\bibitem{candes2011robust}
E.~J. Cand{\`e}s, X.~Li, Y.~Ma, and J.~Wright, ``Robust principal component
  analysis?'' \emph{Journal of the ACM (JACM)}, vol.~58, no.~3, pp. 1--11,
  2011.

\bibitem{Qianqianl2pnorm}
Q.~Wang, Q.~Gao, X.~Gao, and F.~Nie, ``${L}_{2,p}$-norm based {PCA} for image
  recognition,'' \emph{IEEE Trans. Image Process.}, vol.~27, no.~3, pp.
  1336--1346, 2018.

\bibitem{rahmani2016coherence}
M.~Rahmani and G.~K. Atia, ``Coherence pursuit: fast, simple, and robust
  principal component analysis,'' \emph{IEEE Trans. Signal Process.}, vol.~65,
  no.~23, pp. 6260--6275, 2016.

\bibitem{inoue2006equivalence}
K.~Inoue and K.~Urahama, ``Equivalence of non-iterative algorithms for
  simultaneous low rank approximations of matrices,'' in \emph{Proc. IEEE
  Comput. Soc. Conf. Comput. Vis. Pattern Recog.}, 2006, pp. 154--159.

\bibitem{ye2005two}
J.~Ye, R.~Janardan, and Q.~Li, ``Two-dimensional linear discriminant
  analysis,'' in \emph{Advances in Neural Inform. Process. Syst.}, 2005, pp.
  1569--1576.

\bibitem{gu2012low}
Z.~Gu, W.~Lin, B.-S. Lee, and C.~Lau, ``Low-complexity video coding based on
  two-dimensional singular value decomposition,'' \emph{IEEE Trans. Image
  Process.}, vol.~21, no.~2, pp. 674--687, 2012.

\bibitem{hou20172d}
C.~Hou, Y.~Jiao, F.~Nie, T.~Luo, and Z.-H. Zhou, ``2{D} feature selection by
  sparse matrix regression,'' \emph{IEEE Trans. Image Process.}, vol.~26,
  no.~9, pp. 4255--4268, 2017.

\bibitem{shikkenawis20162d}
G.~Shikkenawis and S.~K. Mitra, ``2{D} orthogonal locality preserving
  projection for image denoising,'' \emph{IEEE Trans. Image Process.}, vol.~25,
  no.~1, pp. 262--273, 2016.

\bibitem{li2010l1}
X.~Li, Y.~Pang, and Y.~Yuan, ``${L}_1$-norm-based 2{DPCA},'' \emph{IEEE Trans.
  Syst., Man, Cybern., B Cybern.}, vol.~40, no.~4, pp. 1170--1175, 2010.

\bibitem{yang2004two}
J.~Yang, D.~Zhang, A.~F. Frangi, and J.~Yang, ``Two-dimensional {PCA}: a new
  approach to appearance-based face representation and recognition,''
  \emph{IEEE Trans. Pattern Anal. Mach. Intell.}, vol.~26, no.~1, pp. 131--137,
  2004.

\bibitem{cai2005subspace}
D.~Cai, X.~He, and J.~Han, ``Subspace learning based on tensor analysis,''
  Comput. Sci. Dept., UIUC, Tech. Rep. UIUCDCS-R-2005–2572, 2005.

\bibitem{ye2005generalized}
J.~Ye, ``Generalized low rank approximations of matrices,'' \emph{Mach.
  Learn.}, vol.~61, no. 1-3, pp. 167--191, 2005.

\bibitem{ding20052}
C.~Ding and J.~Ye, ``2-{D}imensional singular value decomposition for 2{D} maps
  and images,'' in \emph{SIAM Intl. Conf. Data Mining,}, 2005, pp. 32--43.

\bibitem{ke2005robust}
Q.~Ke and T.~Kanade, ``Robust ${L}_1$ norm factorization in the presence of
  outliers and missing data by alternative convex programming,'' in \emph{Proc.
  Intl. Conf. Comput. Vis. Pattern Recogn.}, vol.~1, 2005, pp. 739--746.

\bibitem{zhong2013linear}
F.~Zhong and J.~Zhang, ``Linear discriminant analysis based on ${L}_1$-norm
  maximization,'' \emph{IEEE Trans. Image Process.}, vol.~22, no.~8, pp.
  3018--3027, 2013.

\bibitem{liu2017non}
Y.~Liu, Q.~Gao, S.~Miao, X.~Gao, F.~Nie, and Y.~Li, ``A non-greedy algorithm
  for ${L}_1$-norm {LDA},'' \emph{IEEE Trans. Image Process.}, vol.~26, no.~2,
  pp. 684--695, 2017.

\bibitem{huang2008robust}
H.~Huang and C.~Ding, ``Robust tensor factorization using ${R}_1$ norm,'' in
  \emph{Proc. Intl. Conf. Comput. Vis. Pattern Recogn.}, 2008, pp. 1--8.

\bibitem{he2011robust}
R.~He, B.-G. Hu, W.-S. Zheng, and X.-W. Kong, ``Robust principal component
  analysis based on maximum correntropy criterion,'' \emph{IEEE Trans. Image
  Process.}, vol.~20, no.~6, pp. 1485--1494, 2011.

\bibitem{he2011maximum}
R.~He, W.-S. Zheng, and B.-G. Hu, ``Maximum correntropy criterion for robust
  face recognition,'' \emph{IEEE Trans. Pattern Anal. Mach. Intell.}, vol.~33,
  no.~8, pp. 1561--1576, 2011.

\bibitem{liu2007correntropy}
W.~Liu, P.~P. Pokharel, and J.~C. Pr{\'\i}ncipe, ``Correntropy: Properties and
  applications in non-{G}aussian signal processing,'' \emph{IEEE Trans. Signal
  Process.}, vol.~55, no.~11, pp. 5286--5298, 2007.

\bibitem{principe2000information}
J.~C. Pr{\'\i}ncipe, D.~Xu, and J.~Fisher, ``Information theoretic learning,''
  \emph{Unsupervised Adaptive Filtering}, vol.~1, pp. 265--319, 2000.

\bibitem{santamaria2006generalized}
I.~Santamar{\'\i}a, P.~P. Pokharel, and J.~C. Principe, ``Generalized
  correlation function: definition, properties, and application to blind
  equalization,'' \emph{IEEE Trans. Signal Process.}, vol.~54, no.~6, pp.
  2187--2197, 2006.

\bibitem{chen2016generalized}
B.~Chen, L.~Xing, H.~Zhao, N.~Zheng, and J.~C. Pr{\'\i}ncipe, ``Generalized
  correntropy for robust adaptive filtering,'' \emph{IEEE Trans. Signal
  Process.}, vol.~64, no.~13, pp. 3376--3387, 2016.

\bibitem{chen2017robust}
B.~Chen, L.~Xing, X.~Wang, J.~Qin, and N.~Zheng, ``Robust learning with kernel
  mean $p$-power error loss,'' \emph{IEEE Trans. Cyber.}, 2017.

\bibitem{zhao2017kernel}
J.~Zhao and H.~Zhang, ``Kernel recursive generalized maximum correntropy,''
  \emph{IEEE Signal Process. Lett.}, vol.~24, no.~12, pp. 1832--1836, 2017.

\bibitem{erdogmus2002generalized}
D.~Erdogmus and J.~C. Principe, ``Generalized information potential criterion
  for adaptive system training,'' \emph{IEEE Trans. Neural Networks}, vol.~13,
  no.~5, pp. 1035--1044, 2002.

\bibitem{hild2006feature}
K.~E. Hild, D.~Erdogmus, K.~Torkkola, and J.~C. Principe, ``Feature extraction
  using information-theoretic learning,'' \emph{IEEE Trans. Pattern Anal. Mach.
  Intell.}, vol.~28, no.~9, pp. 1385--1392, 2006.

\bibitem{huber1981robust}
P.~J. Huber, ``Robust {S}tatistics,'' Wiley, 1981.

\bibitem{gao2007center}
Q.-B. Gao and Z.-Z. Wang, ``Center-based nearest neighbor classifier,''
  \emph{Pattern Recogn.}, vol.~40, no.~1, pp. 346--349, 2007.

\bibitem{ding2008tensor}
C.~Ding, H.~Huang, and D.~Luo, ``Tensor reduction error analysis-- applications
  to video compression and classification,'' in \emph{Proc. Intl. Conf. Comput.
  Vis. Pattern Recogn.}, 2008, pp. 1--8.

\bibitem{huang2008simultaneous}
H.~Huang, C.~Ding, D.~Luo, and T.~Li, ``Simultaneous tensor subspace selection
  and clustering: the equivalence of high order {SVD} and {K}-means
  clustering,'' in \emph{Proc. Int. Conf. Knowl. Disc. Data Min. (KDD)}, 2008,
  pp. 327--335.

\bibitem{CHH05}
D.~Cai, X.~He, and J.~Han, ``Document clustering using locality preserving
  indexing,'' \emph{IEEE Trans. Knowledge and Data Eng.}, vol.~17, no.~12, pp.
  1624--1637, 2005.

\bibitem{lovasz2009matching}
L.~Lov{\'a}sz and M.~D. Plummer, \emph{Matching Theory}.\hskip 1em plus 0.5em
  minus 0.4em\relax American Mathematical Soc., 2009, vol. 367.

\bibitem{strehl2002cluster}
A.~Strehl and J.~Ghosh, ``Cluster ensembles--a knowledge reuse framework for
  combining multiple partitions,'' \emph{Journal of Machine Learning Research},
  vol.~3, no. Dec, pp. 583--617, 2002.

\bibitem{zhang2017robust}
Z.~Zhang, F.~Li, M.~Zhao, L.~Zhang, and S.~Yan, ``Robust neighborhood
  preserving projection by nuclear/$l_{2, 1}$-norm regularization for image
  feature extraction,'' \emph{IEEE Trans. Image Process.}, vol.~26, no.~4, pp.
  1607--1622, 2017.

\bibitem{ding2004k}
C.~Ding and X.~He, ``K-means clustering via principal component analysis,'' in
  \emph{Proc. 21st Int. Conf. Mach. Learning}, 2004, pp. 225--232.

\bibitem{rodriguez2014clustering}
A.~Rodriguez and A.~Laio, ``Clustering by fast search and find of density
  peaks,'' \emph{Science}, vol. 344, no. 6191, pp. 1492--1496, 2014.

\end{thebibliography}

\end{document}